%% file: final.tex
\documentclass[10pt,twocolumn,letterpaper]{article}

\usepackage{arxiv}
\usepackage{times}
\usepackage{epsfig}
\usepackage{graphicx}
\usepackage{amsmath}
\usepackage{amssymb}
\usepackage{float}
\usepackage{comment}
\usepackage{caption}
\usepackage{stfloats}
\usepackage{amsfonts}
\usepackage[toc,page]{appendix}

\usepackage[breaklinks=true,bookmarks=false]{hyperref}

\arxivfinalcopy 

\def\arxivPaperID{****} 

\newcommand{\para}{\vspace*{-3mm}\paragraph}
\newcommand{\jw}[1]{#1}

\newcommand{\OURS}{TextureNet}

\begin{document}

\title{\OURS: Consistent Local Parametrizations for Learning from High-Resolution Signals on Meshes}

\author{
Jingwei Huang$^{1}$ \qquad Haotian Zhang$^{1}$  \qquad Li Yi$^{1}$ \qquad Thomas Funkhouser$^{2}$
\vspace{0.1cm} \\ 
Matthias Nie{\ss}ner$^{3}$ \qquad Leonidas Guibas$^{1}$
\vspace{0.2cm} \\ 
$^{1}$Stanford University \qquad $^{2}$Princeton University \qquad $^{3}$Technical University of Munich
\vspace{-0.3cm}
}
\twocolumn[{%
	\renewcommand\twocolumn[1][]{#1}%
	\maketitle
	\begin{center}
		\vspace{-0.4cm}
		\includegraphics[width=\linewidth]{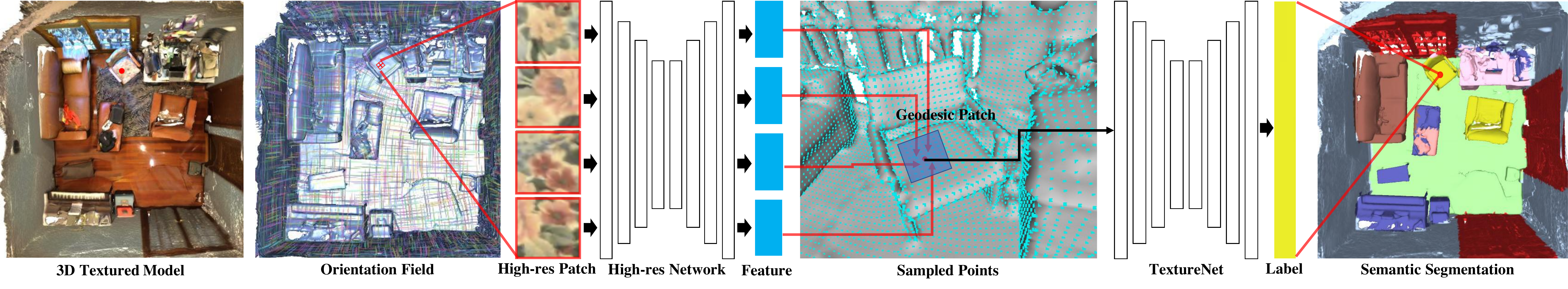}
		\vspace{-0.4cm}
		\captionof{figure}{\OURS{} takes as input a 3D textured mesh.  The mesh is parameterized with a consistent 4-way rotationally symmetric (4-RoSy) field, which is used to extract oriented patches from the texture at a set of sample points.   Networks of 4-RoSy convolutional operators extract features from the patches and used for 3D semantic segmentation.}
		\label{fig:teaser}
	\end{center}    
}]

\maketitle

\input{0Abstract}
\input{1Introduction}
\input{2RelatedWorks}

\input{3Approach}
\input{4Evaluation}
\input{5Conclusion}

\section*{Acknowledgements}
This work is supported in part by Google, Intel, Amozon, a Vannevar Bush faculty fellowship, a TUM Foundation Fellowship, a TUM-IAS Rudolf M{\"o}{\ss}bauer Fellowship, the ERC Starting Grant \emph{Scan2CAD}, and the NSF grants VEC-1539014/1539099, CHS-1528025 and IIS-1763268.  It makes use of data from Matterport.

{\small
\bibliographystyle{ieee}.
\bibliography{egbib}
}

\clearpage

\input{6Appendix}
\end{document}

%% file: 0Abstract.tex
\begin{abstract}
\vspace{-0.1cm}
We introduce, \OURS{}, a neural network architecture designed to extract features from high-resolution signals associated with 3D surface meshes (e.g., color texture maps).  The key idea is to utilize a 4-rotational symmetric (\textit{4-RoSy}) field to define a domain for convolution on a surface.  Though 4-RoSy fields have several properties favorable for convolution on surfaces (low distortion, few singularities, consistent parameterization, etc.), orientations are ambiguous up to 4-fold rotation at any sample point.  So, we introduce a new convolutional operator invariant to the 4-RoSy ambiguity and use it in a network to extract features from high-resolution signals on geodesic neighborhoods of a surface. In comparison to alternatives, such as PointNet-based methods which lack a notion of orientation, the coherent structure given by these neighborhoods results in significantly stronger features.  As an example application, we demonstrate the benefits of our architecture for 3D semantic segmentation of textured 3D meshes.  The results show that our method outperforms all existing methods on the basis of mean IoU by a significant margin in both geometry-only (6.4\%) and RGB+Geometry (6.9-8.2\%) settings. 
\end{abstract}

%% file: 1Introduction.tex
\vspace{-0.1in}
\section{Introduction}

In recent years, there has been tremendous progress in RGB-D scanning methods that allow reliable tracking and reconstruction of 3D surfaces using hand-held, consumer-grade devices \cite{curless1996volumetric,izadi2011kinectfusion,newcombe2011kinectfusion,niessner2013real,whelan2016elasticfusion,kahler2015very,dai2017bundlefusion}.  Though these methods are now able to reconstruct high-resolution textured 3D meshes suitable for visualization, understanding the 3D semantics of the scanned scenes is still a relatively open research problem.

There has been a lot of recent work on semantic segmentation of 3D data using convolutional neural networks (CNNs).  Typically, features extracted from the scanned inputs (e.g., positions, normals, height above ground, colors, etc.) are projected onto a coarse sampling of 3D locations, and then a network of 3D convolutional filters is trained to extract features for semantic classification -- e.g., using convolutions over voxels \cite{wu20153d,maturana2015voxnet,qi2016volumetric,song2017semantic,dai2017scannet,dai2018scancomplete}, octrees \cite{riegler2017octnet}, point clouds \cite{qi2017pointnet,qi2017pointnet++}, or mesh vertices \cite{masci2015geodesic}.  The advantage of this approach over 2D image-based methods is that convolutions operate directly on 3D data, and thus are relatively unaffected by view-dependent effects of images, such as perspective, occlusion, lighting, and background clutter.   However, the resolution of current 3D representations is generally quite low (2cm is typical), and so the ability of 3D CNNs to discriminate fine-scale semantic patterns is usually far below their color image counterparts \cite{long2015fully,he2017mask}.

To address this issue, we propose a new convolutional neural network, \OURS, that extracts features directly from high-resolution signals associated with 3D surface meshes.  Given a map that associates high-resolution signals with a 3D mesh surface (e.g., RGB photographic texture), we define convolutional filters that operate on those signals within domains defined by geodesic surface neighborhoods.   This approach combines the advantages of feature extraction from high-resolution signals (as in \cite{dai20183dmv}) with the advantages of view-independent convolution on 3D surface domains (as in \cite{tatarchenko2018tangent}).   This combination is important for the example in labeling the chair in Figure \ref{fig:teaser}, whose surface fabric is easily recognizable in a color texture map.

During our investigation of this approach, we had to address several research issues, the most significant of which is how to define on geodesic neighborhoods of a mesh.   One approach could be to compute a global UV parameterization for the entire surface and then define convolutional operators directly in UV space; however, that approach would induce significant deformation due to flattening, not always follow surface features, and/or produce seams at surface cuts.  Another approach could be to compute UV parameterizations for local neighborhoods independently; however, then adjacent neighborhoods might not be oriented consistently, reducing the ability of a network to learn orientation-dependent features.   Instead, we compute a 4-RoSy (four-fold rotationally symmetric) field on the surface using QuadriFlow~\cite{huang2018quadriflow} and define a new 4-RoSy convolutional operator that explicitly accounts for the 4-fold rotational ambiguity of the cross field parameterization. \jw{Here, 4-RoSy field is a set of tangent directions associated with vertices, where neighboring directions are parallel to each other by rotating one of them around surface normal by 360K/4 degrees} ($K\in \mathbb{Z}$).  Since the 4-RoSy field from QuadriFlow has no seams, aligns to shape features, induces relatively little distortion, has few singularities, and consistently orients adjacent neighborhoods (up to 4-way rotation), it provides a favorable trade-off between distortion and orientation invariance.

Results on 3D semantic segmentation benchmarks show an improvement of 4-RoSy convolution on surfaces over alternative geometry-only approaches (by 6.4\%), plus significantly further improvement when applied to high-resolution color signals (by 6.9-8.2\% ).  With ablation studies, we verify the importance of the consistent orientation of a 4-RoSy field and demonstrate that our sampling and convolution operator works better than other alternatives.  

Overall, our core research contributions are:
 \begin{itemize}
 \setlength{\topsep}{0pt}
 \setlength{\parsep}{0pt}
 \setlength{\partopsep}{0pt}
 \setlength{\parskip}{0pt}
 \setlength{\itemsep}{2pt}
 \vspace*{-2mm}
     \item a novel learning-based method for extracting features from high-resolution signals living on surfaces embedded in 3D, based on consistent local parameterizations,
     \item a new 4-RoSy convolutional operator designed for cross fields on general surfaces in 3D,
     \item a new deep network architecture, \OURS, composed of 4-RoSy convolutional operators,
     \item an extensive experimental investigation of alternative convolutional operators for semantic segmentation of surfaces in 3D.
 \end{itemize}

%% file: 2RelatedWorks.tex
\begin{figure*}
\includegraphics[width=\linewidth]{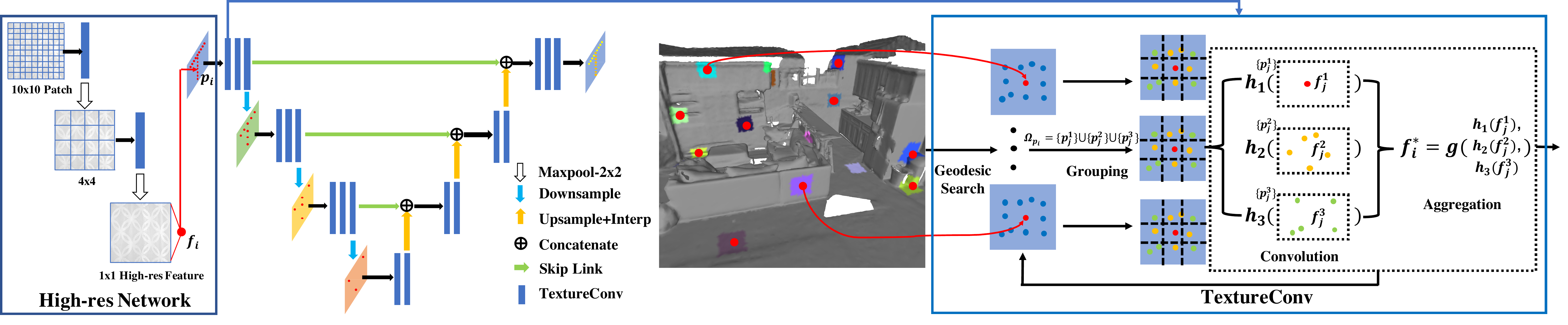}
\vspace{-0.7cm}
\caption{\OURS{} architecture. We propose a UNet~\cite{ronneberger2015u} architecture for hierarchical feature extraction. The key innovation in the architecture is the texture convolution layer. We efficiently query the local geodesic patch for each surface point, associate each neighborhood with a local, orientation-consistent texture coordinate. This allows us to extract the local 3D surface features as well as high-resolution signals such as associated RGB input.}
\label{fig:approach-network}
\vspace{-0.1in}
\end{figure*}

\vspace{-0.1in}
\section{Related Work}

\paragraph{3D Deep Learning.}
With the availability of 3D shape databases \cite{wu20153d,chang2015shapenet,song2017semantic} and real-world labeled 3D scanning data \cite{song2015sun,armeni2017joint,dai2017scannet,chang2017matterport3d}, there is significant interest in deep learning on three-dimensional data.   Early work developed CNNs operating on 3D volumetric grids \cite{wu20153d,maturana2015voxnet}.  They have been used for 3D shape classification  \cite{qi2016volumetric,riegler2017octnet}, semantic  segmentation \cite{dai2017scannet,dai2018scancomplete}, object completion \cite{dai2017shape}, and scene completion \cite{dai2018scancomplete}.   More recently, researchers have developed methods that can take a 3D point cloud as input to a neural network and predict object classes or semantic point labels \cite{qi2017pointnet,qi2017pointnet++,tatarchenko2018tangent,su2018splatnet,atzmon2018point}.  AtlasNet~\cite{groueix2018papier} learns to generate surfaces of the 3D shape.  In our work, we utilize a sparse point sampled data representation, however, we exploit high resolution signals on geometric surface structures with a new 4-RoSy surface convolution kernel.

\para{Convolutions on Meshes.}
Several researchers have proposed methods for applying convolutional neural networks intrinsically on manifold meshes.  FeaStNet~\cite{verma2018feastnet} proposes a graph operator that establishes correspondences between filter weights. Jiang \textit{et al.}~\cite{jiang2019spherical} applies differential operators on unstructured spherical grids.
GCNN~\cite{masci2015geodesic} proposes using discrete patch operators on tangent planes parameterized by radius and angles. 
However, the orientation of their selected geodesic patches is arbitrary, and the parameterization is highly distorted or inconsistent at regions with high Gaussian curvature. 
ACNN~\cite{boscaini2016learning} observes this limitation and introduces the anisotropic heat kernels derived from principal curvatures. MoNet~\cite{monti2017geometric} further generalizes the architecture with the learnable gaussian kernels for convolutions.
The principal curvature based frame selection method is adopted by Xu \textit{et al.}~\cite{xu2017directionally} for segmentation of nonrigid surfaces, by Tatarchenko \textit{et al.}~\cite{tatarchenko2018tangent} for semantic segmentation of point clouds, and by ADD~\cite{boscaini2016anisotropic} for shape correspondence in the spectral domain. 
It naturally removes orientation ambiguity but fails to consider frame inconsistency problem, which is critical when performing feature aggregation.  Its problems are particularly pronounced in indoor scenes (which often have many planar regions where principal curvature is undetermined) and in real-world scans (which often have noisy and uneven sampling where consistent principal curvatures are difficult to predict).   In contrast, we define a 4-RoSy field that provides consistent orientations for neighboring convolution domains.

\para{Multi-view and 2D-3D Joint Learning.}
Other researchers have investigated how to incorporate features from RGB inputs to 3D deep networks.  The typical approach is to simply assign color values to voxels, points, or mesh vertices and treat them as additional feature channels.
However, given that geometry and RGB data are at vastly different resolutions, this approach leads to significant downsampling of the color signal and thus does not take full advantage of the high-frequency patterns therein.   An alternative approach is to combine features extracted from RGB images in a multi-view CNN \cite{su2015multi}. This approach has been used for 3D semantic segmentation in 3DMV \cite{dai20183dmv}, where features are extracted from 2D RGB images and then back-projected into a 3D voxel grid where they are merged and further processed with 3D voxel convolutions.  Like our approach, 3DMV processes high-resolution RGB signals; however it convolves them in a 2D image plane, where occlusions and background clutter are confounding.  In contrast, our method directly convolves high-resolution signals intrinsically on the 3D surface which is view-independent.

%% file: 3Approach.tex
\section{Approach}
Our approach performs convolutions on high-resolution signals with geodesic convolutions directly on 3D surface meshes.
The input is a 3D mesh associated with a high-resolution surface signal (e.g., a color texture map), and the outputs are learned features for a dense set of sample points that can be used for semantic segmentation and other tasks.   

Our main contribution is defining a smooth, consistently oriented domain for surface convolutions based on four-way rotationally symmetric (4-RoSy) fields.   We observe that 3D surfaces can be mapped with low-distortion to two-dimensional parameterizations anchored at dense sample points with locally consistent orientations and few singularities if we allow for a four-way ambiguity in the orientation at the sample points.   We leverage that observation in \OURS{} by computing a 4-RoSy field and point sampling using QuadriFlow~\cite{huang2018quadriflow} and then building a network using new 4-RoSy convolutional filters (TextureConv) that are invariant to the four-way rotational ambiguity.   

We utilize this network design to learn and extract features from high-resolution signals on surfaces by extracting surface patches with high-resolution signals oriented by the 4-RoSy field at each sample point.   The surface patches are convolved by a few TextureConv layers, pooled at sample points, and then convolved further with TextureConv layers in a UNet~\cite{ronneberger2015u} architecture, as shown in figure~\ref{fig:approach-network}.  For down-sampling and up-sampling, we use the furthest point sampling and three-nearest neighbor interpolation method proposed by PointNet++~\cite{qi2017pointnet++}.  The output of the network is a set of features associated with point samples that can be used for classification and other tasks.   The following sections describe the main components of the network in detail.

\subsection{High-Resolution Signal Representation}
\label{sec:high-res}
Our network takes as input a high-resolution signal associated with a 3D surface mesh.   In the first steps of processing, it generates a set of sample points on the mesh and defines a parameterized high-resolution patch for each sample (Section \ref{sec:approach-param}) \jw{as follow}: For each sample point $\mathbf{p}_i$, we first compute its geodesic neighborhood $\Omega_{\rho}(\mathbf{p}_i)$ (\jw{Eq.}~\ref{eq:omega}) with radius $\rho$. Then, we sample an NxN point cloud $\{\mathbf{q}_{xy}|-N/2\leq x,y<N/2\}$. The texture coordinate for $\mathbf{q}_{xy}$ is $((x+0.5)d,(y+0.5)d)$ -- $d$ is the distance between the adjacent pixels in the texture patch. In practice, we select $N=10$ and $d=4$mm. Finally, we use our newly proposed ``TextureConv'' and max-pooling operators (Section \ref{sec:approach-conv}) to extract the high-res feature $\mathbf{f}_i$ for each point $\mathbf{p}_i$.

\subsection{4-RoSy Surface parameterization}
\label{sec:approach-param}
\begin{figure}
    \centering
    \includegraphics[width=\linewidth]{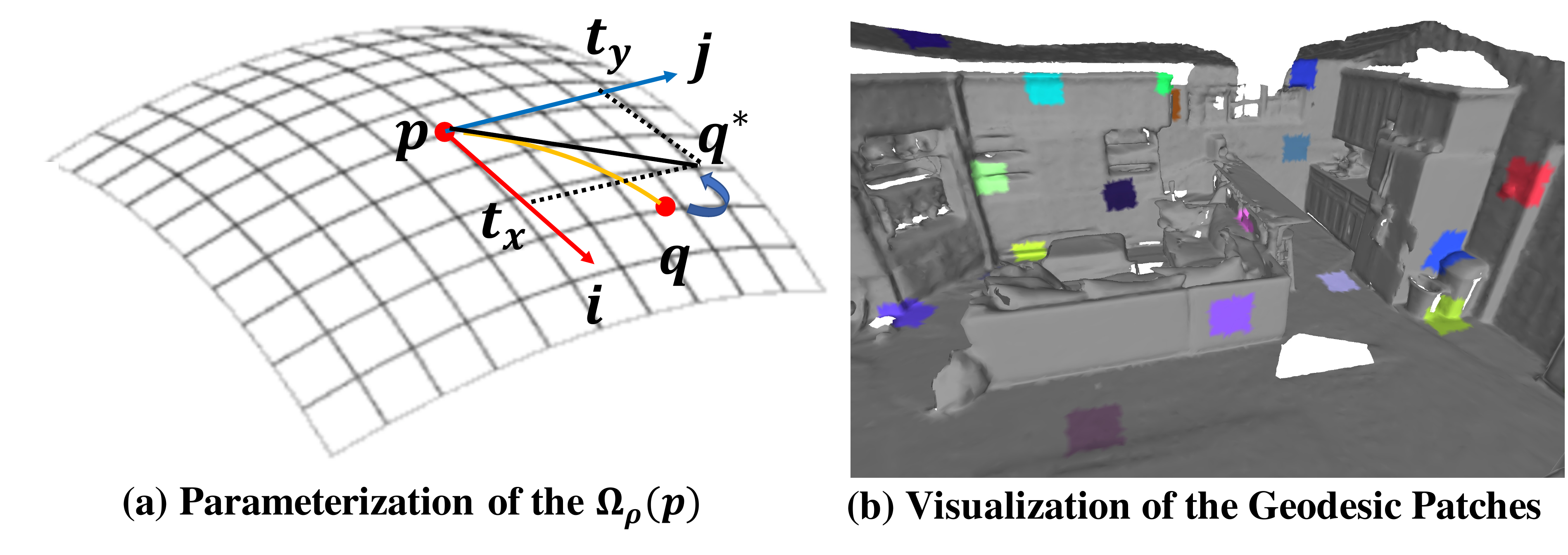}
    \vspace{-0.8cm}
    \caption{(a) Local texture coordinate. (b) Visualization of geodesic neighborhoods $\Omega_\rho$ ($\rho$ = 20 cm) of a set of randomly sampled vertices.}
    \vspace{-0.3cm}
    \label{fig:geodesic}
\end{figure}

\begin{figure}
     \includegraphics[width=\linewidth]{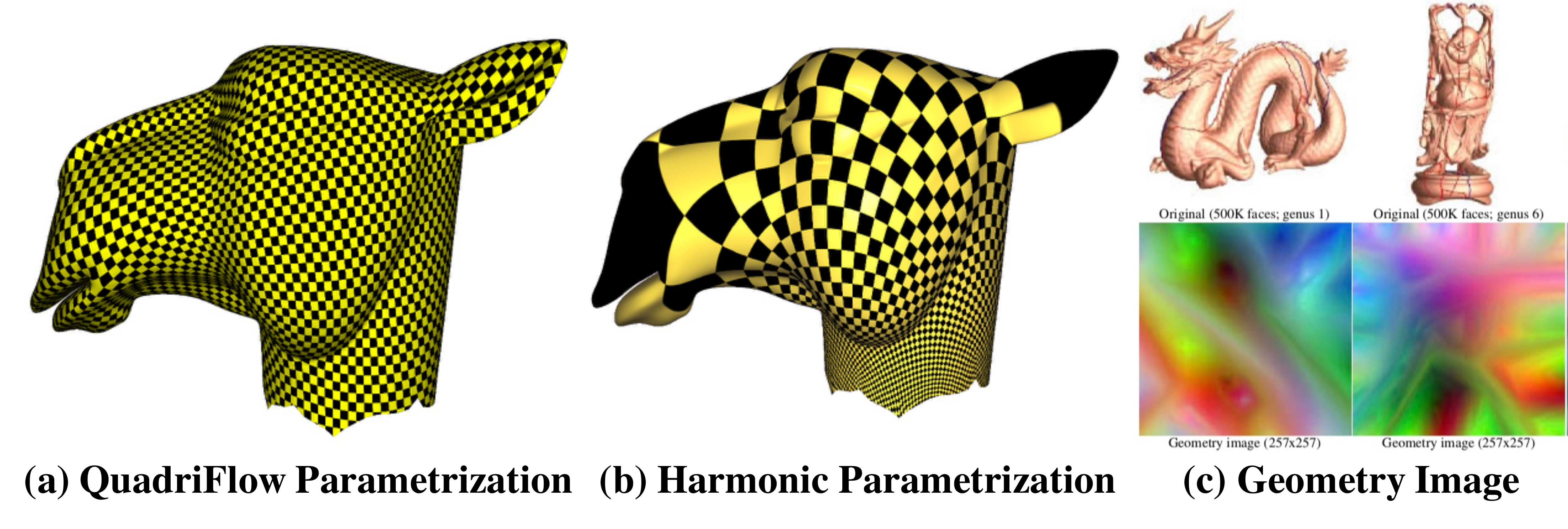}
     \vspace{-0.8cm}
     \caption{(a) With appropriate method like Quadriflow, we can get the surface parameterization aligning to shape features with negligible distortions. (b) Harmonic parameterizations leads to high distortion in the scale. (c) Geometry images~\cite{gu2002geometry} result in high distortion in the orientation.}
    \vspace{-0.15in}
     \label{fig:param}
 \end{figure}

 A critical aspect of our network is to define a consistently-oriented geodesic surface parameterization for any position on a 3D mesh. Starting with some basic definitions, for a sampled point $\mathbf{p}$ on the surface, we can locally parameterize its tangent plane by two orthogonal tangent vectors $\mathbf{i}$ and $\mathbf{j}$.  Also, for any point $\mathbf{q}$ on the surface, there exists a shortest path on the surface connecting $\mathbf{p}$ and $\mathbf{q}$, e.g., the orange path in figure~\ref{fig:geodesic}(a). By unfolding it to the tangent plane, we can map $\mathbf{q}$ along the shortest path to $\mathbf{q^*}$.   Using these constructs, we define the local texture coordinate $\mathbf{q}$ in $\mathbf{p}$'s neighborhood as
 \begin{equation*}
     \mathbf{t}_{\mathbf{p}}(\mathbf{q}) = \begin{bmatrix}
     \mathbf{i}^T & \mathbf{j}^T
     \end{bmatrix}(\mathbf{q}^*-\mathbf{p}).
 \end{equation*}
 We additionally define the local geodesic neighborhood of $\mathbf{p}$ with receptive field $\rho$ as
\begin{equation}
\Omega_\rho(\mathbf{p}) = \{\mathbf{q}\;|\;||\mathbf{t}_{\mathbf{p}}(\mathbf{q})||_\infty < \rho\}.
\label{eq:omega}
\end{equation}
The selection for the set of mesh sampled positions $\{\mathbf{p}\}$ and their tangent vectors $\mathbf{i}$ and $\mathbf{j}$ is critical for the success of learning on a surface domain.  Ideally, we would select points whose spacing is uniform and whose tangent directions are consistently oriented at neighbors, such that the underlying parameterization has no distortions or seams, as shown in Figure~\ref{fig:param}(a).  With those properties, we could learn convolutional operators with translation invariance exactly as we would for images.  Unfortunately, these properties are only achievable if the surface is a flat plane.   
For a general 3D surface, we can only hope to select a set of point samples and tangent vectors that minimize deviations between spacings of points and distortions of local surface parameterizations. Figure~\ref{fig:param}(b) shows an example where harmonic surface parameterization introduces large-scale distortion -- a 2D convolution would include a large receptive field at the nose but a small one at the neck. Figure~\ref{fig:param}(c) shows a geometry image~\cite{gu2002geometry} parameterization with high distortion in the orientation -- convolutions on such a map would have randomly distorted and irregular receptive fields, making it difficult for a network to learn canonical features.
 
Unfortunately, a smoothly varying direction field on the surface is usually hard to obtain. According to the study of the direction field design~\cite{ray2008n,lai2010metric}, the best-known approach to mitigate the distortion is to compute a four-way rotationally symmetric (\emph{4-RoSy}) \emph{orientation field}, which minimizes the deviation by incorporating directional ambiguity. Additionally, the orientation field needs a consistent definition among different geometries, and the most intuitive way is to make it align with the shape features like the principal curvatures. Fortunately, the extrinsic energy is used by \cite{jakob2015instant,huang2018quadriflow} to realize it. Therefore, we compute the extrinsic 4-Rosy orientation field at a uniform distribution of point samples using QuadriFlow~\cite{huang2018quadriflow} and use it to define the tangent vectors at any position on the surface. Because of the directional ambiguity, we randomly pick one direction from the cross as $\mathbf{i}$ and compute $\mathbf{j}=\mathbf{n}\times \mathbf{i}$ for any position. 

\begin{figure}
    \centering
    \includegraphics[width=\linewidth]{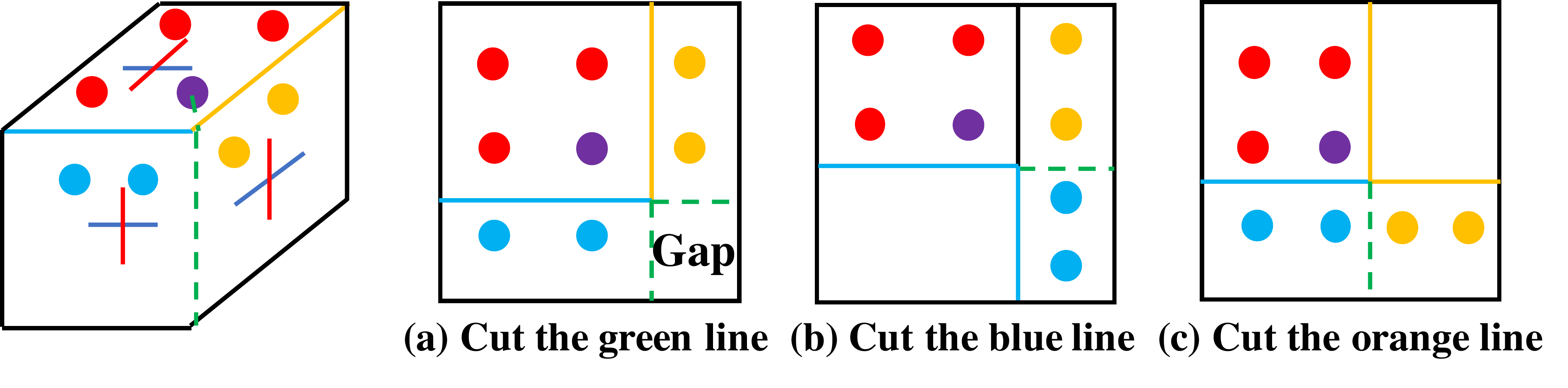}
    \vspace{-0.8cm}
    \caption{At the singularity of the cube, (a)-(c) provides three different ways of unfolding the local neighborhood. Such ambiguity is removed around the singularity by our texture coordinate definition using the shortest path. For the purple point, (a) is a valid neighborhood, while the blue points in (b) and orange points in (c) are unfolded along the paths which are not the shortest. Similarly, the ambiguity of the gap location is removed.}
    \vspace{-0.15in}
    \label{fig:wrap}
\end{figure}

Although there is a 4-way rotational ambiguity in this local parameterization of the surface (which will be addressed with a new convolutional operator in the next section), the resulting 4-RoSy field provides a way to extract geodesic neighborhoods consistently across the entire surface, even near singularities. 
Figure~\ref{fig:wrap} (a,b,c) shows the ambiguity of possible unfolded neighborhoods at a singularity.  Since QuadriFlow~\cite{huang2018quadriflow} treats singularities as faces rather than vertices, all sampled positions have the well-defined orientation field. More importantly, the parameterization of every geodesic neighborhood is well-defined with our shortest path patch parameterization. For example, only Figure~\ref{fig:wrap}(a) is a valid parameterization for the purple spot, while the location for the blue and orange spots in Figures~\ref{fig:wrap}(b) and (c) are unfolded along the paths that are not the shortest. Unfolding a geodesic neighborhood around the singularity also causes another potential issue that a seam cut is usually required, leading to a gap at the 3-singularity or multiple-surface coverage at the 5-singularity. For example, there is a gap at the bottom-right corner in Figure~\ref{fig:wrap}(a) caused by the seam cut shown as the green dot line. Fortunately, the location of the seam is also well-defined with our shortest-path definition: it must be the shortest geodesic path going through the singularity. Therefore, our definition of the local neighborhood guarantees a canonical way of surface parameterization even around corners and singularities.

\subsection{4-RoSy Surface Convolution Operator}
\label{sec:approach-conv}
\begin{figure}
     \centering
     \begin{minipage}{0.32\linewidth}
     \centering
     \includegraphics[width=\linewidth,height=\linewidth]{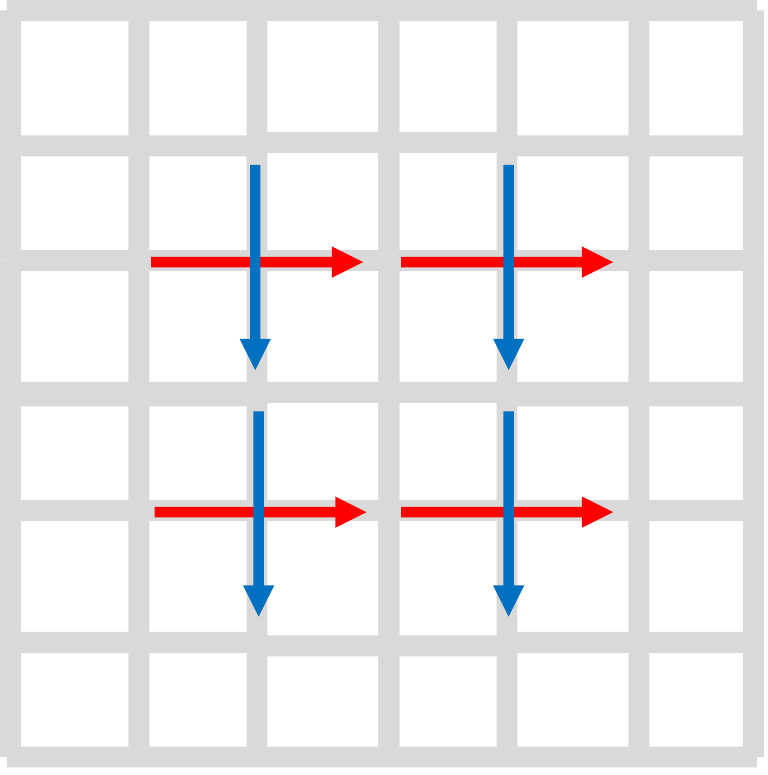}\\
     \footnotesize{
     (a) Image Coordinate
     }
     \end{minipage}
     \begin{minipage}{0.32\linewidth}
     \centering
     \includegraphics[width=\linewidth,height=\linewidth]{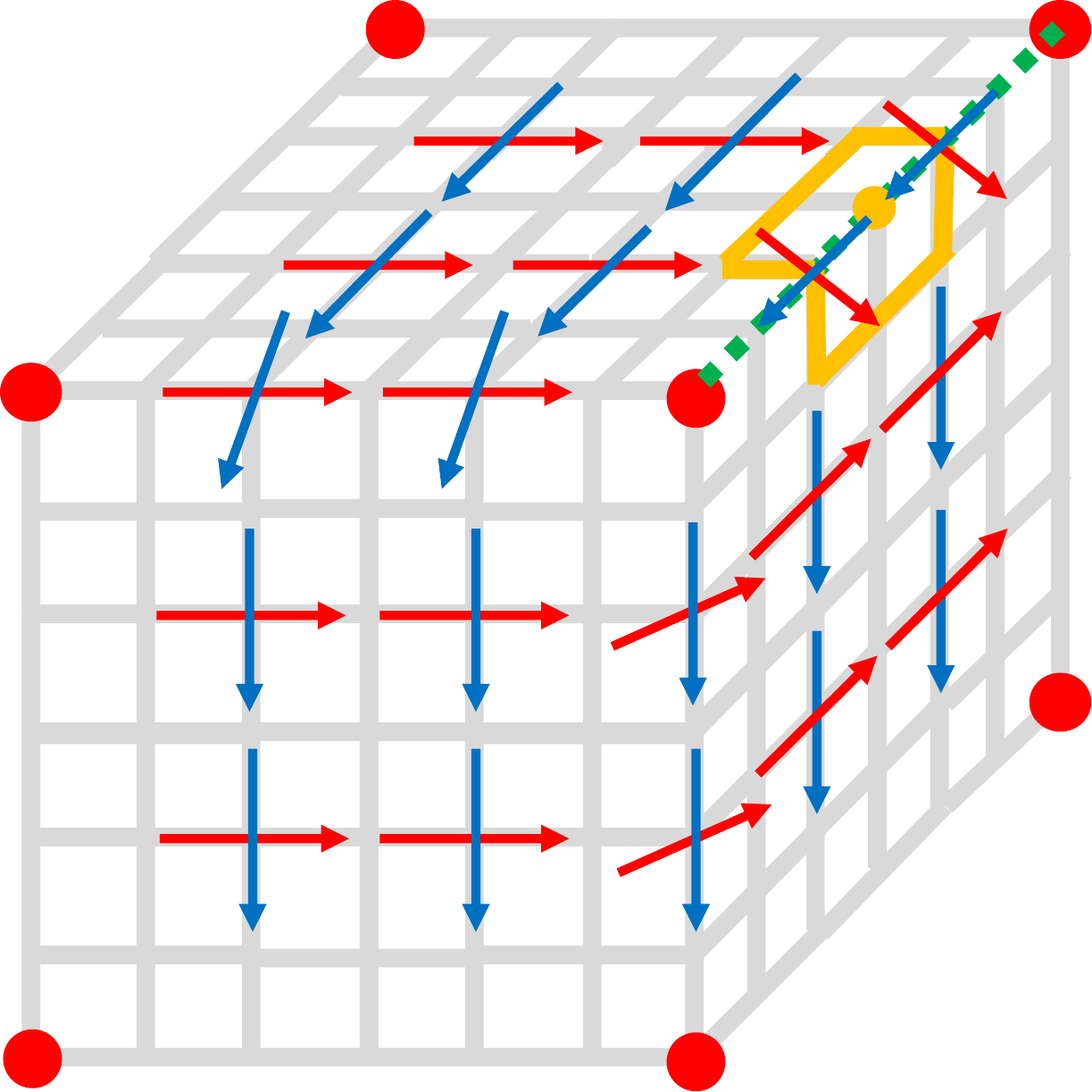}\\
     \footnotesize{
     (b) 3D parametrization
     }
     \end{minipage}
     \begin{minipage}{0.32\linewidth}
     \centering
     \includegraphics[width=\linewidth,height=\linewidth]{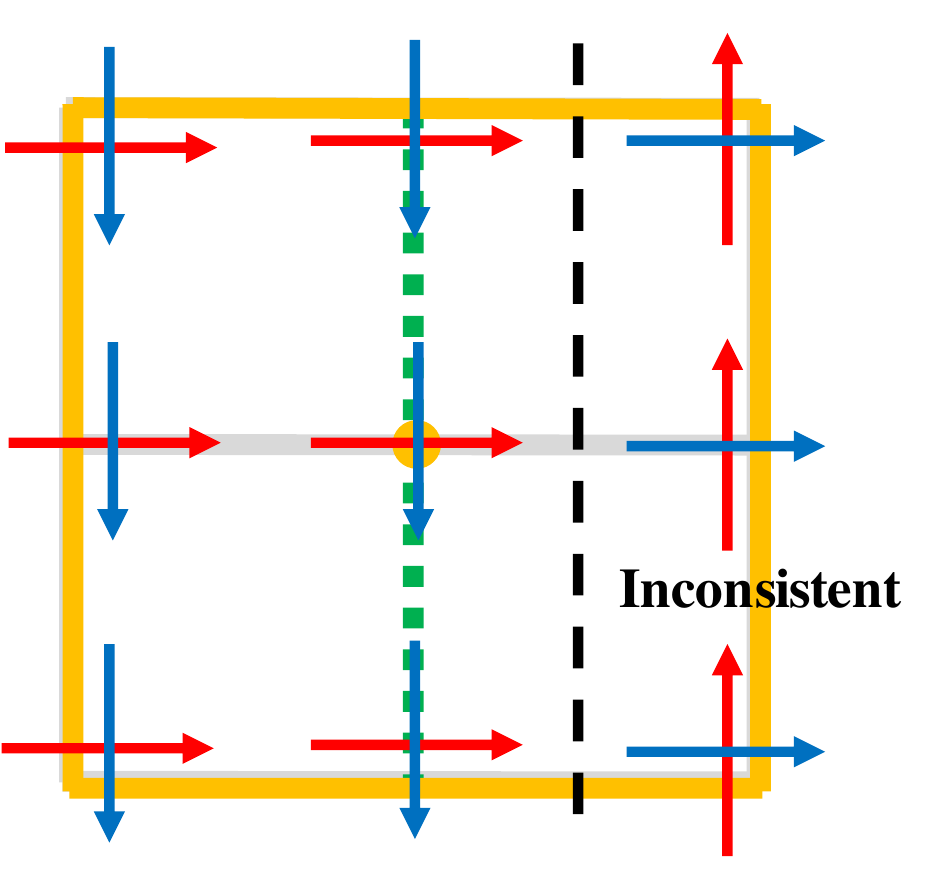}\\
     \footnotesize{
     (c) Inconsistent Frame
     }
     \end{minipage}
     \vspace{-0.2cm}
     \caption{(a) Traditional convolution kernel on a regular grid. (b) Frames defined by the orientation field on a 3D cube. (c) For the patch highlighted in orange in (b), multi-layer feature aggregation would be problematic with traditional convolution due to the frame inconsistency caused by the directional ambiguity of the orientation field.}
    \vspace{-0.2in}
     \label{fig:singular}
 \end{figure}

\OURS~is a network architecture composed of convolutional operators acting on geodesic neighborhoods of sample points with 4-RoSy parameterizations.
The input to each convolutional layer is three-fold: 1) a set of 3D sample points associated with features (e.g., RGB, normals, or features computed from high-resolution surface patches or previous layers); 2) a coordinate system stored as two tangent vectors representing the 4-RoSy cross field for each point sample; and 3) a coarse triangle mesh, where each face is associated with the set of extracted sampled points and connectivity indices that support fast geodesic patch query and texture coordinate computation for the samples inside a geodesic neighborhood, much like the PTex~\cite{burley2008ptex} representation for textures. 

Our key contribution in this section is the design of a convolution operator suitable for 4-RoSy fields. 
The problem is that we cannot use traditional 3x3 convolution kernels on domains parameterized with 4-RoSy fields without inducing inconsistent feature aggregation at higher levels.  Figure~\ref{fig:singular} demonstrates the problem for a simple example.  Figure~\ref{fig:singular}(a) shows 3x3 convolution in a traditional flat domain. Figure~\ref{fig:singular}(b) shows the frames defined by our 4-RoSy orientation field of the 3D cube where red spots represent the singularities. Although the cross-field in the orange patch is consistent under the 4-RoSy metric, the frames are not parallel when they are unfolded into a plane (figure~\ref{fig:singular}(c)). Aggregation of features inside such a patch is therefore problematic.

``TextureConv'' is our solution to remove the directional ambiguity. It consists of four layers (in figure~\ref{fig:approach-network}), including geodesic patch search, texture space grouping, convolution and aggregation. To extract the geodesic patch for each input point $\Omega_\rho(\mathbf{p})$, we use breadth-first search with the priority queue to extract the face set in the order of \jw{geodesic distance from face center to $\mathbf{p}$}. We estimate the texture coordinate at the face center as well as its local tangent coordinate system, recorded as $(\mathbf{t}_f,\mathbf{i}_f,\mathbf{j}_f)$. In order to expand the search tree from face $u$ to $v$, we can approximate the texture coordinate at the face center as $\mathbf{t}_{v} = \mathbf{t}_{u} + (\mathbf{i}_u,\mathbf{j}_u)^T (\mathbf{c}_v - \mathbf{c}_u)$, where $\mathbf{c}_f$ represents the center position of the face $f$. $\mathbf{i}_v$ and $\mathbf{j}_v$ can be computed by rotating the coordinate system around the shared edge from face $u$ to $v$. After having the face set inside the geodesic patch, we can find the sampled points set associated with these faces. We estimate the texture coordinate of every sampled point $\mathbf{q}$ associated with each face $f$ as $\mathbf{t}_\mathbf{p}(\mathbf{q})=\mathbf{t}_f+(\mathbf{i}_f,\mathbf{j}_f)^T(\mathbf{q} - \mathbf{c}_f)$. By testing $||\mathbf{t}_\mathbf{p}(\mathbf{q})||_\infty < \rho$, we can determine the sampled points inside the geodesic patch $\Omega_\rho(\mathbf{p})$.

The texture space grouping layer segments the local neighborhood into 3x3 patches in the texture space, each of which is a square with edge length as $2\rho/3$, as shown in figure~\ref{fig:approach-network} (after the ``grouping arrow''). We could directly borrow the image convolution method linearly transform each point feature with 9 different weights according to their belonging patch. However, we propose a 4-RoSy convolution kernel to deal with the directional ambiguity. As shown in figure~\ref{fig:approach-network}, all sampled points can be categorized as at the corners ($\{\mathbf{p}_j^1\}$), edges ($\{\mathbf{p}_j^2\}$) or the center ($\{\mathbf{p}_j^3\}$). Each sampled point feature is convolved with a 1x1 convolution as $h_1$, $h_2$ or $h_3$ based on its category. The extracted 4-rosy feature removes the ambiguity and allows higher-level feature aggregation. The \jw{channel-wise} aggregation operator $g$ can be max-pooling or average-pooling followed by the ReLu layer. In the task for semantic segmentation, we choose max-pooling since it is better at preserving salient signals.

%% file: 4Evaluation.tex
\section{Evaluation}

To investigate the performance of \OURS{}, we ran a series of 3D semantic segmentation experiments for indoor scenes.   In all experiments, we train and test on the standard splits of the ScanNet~\cite{dai2017scannet} and Matterport3D~\cite{dai2017scannet} datasets.  Following previous works, we report mean class intersection-over-union (mIoU) results for ScanNet and mean class accuracy for Matterport3D.

\para{Comparison to State-of-the-Art.}
\label{sec:eval-result}

\begin{table*}
    \centering
    \tabcolsep=0.05cm
    \begin{tabular}{|c|c|c|c|c|c|c|c|c|c|c|c|c|c|c|c|c|c|c|c|c||c|}
        \hline
        Input & wall & floor & cab & bed & chair & sofa & table & door & wind & shf & pic & cntr & desk & curt & fridg & show & toil & sink & bath & other & avg\\
        \hline
        PN$^+$~\cite{qi2017pointnet++} & 66.4 & 91.5 & 27.8 & 56.3 & 64.0 & 52.7 & 37.3 & 28.3 & 36.1 & 59.2 & 6.7 & 28.0 & 26.2 & 45.4 & 25.6 & 22.0 & 63.5 & 38.8 & 54.4 & 20.0 & 42.5\\
        \hline
        SplatNet~\cite{su2018splatnet} & \textbf{69.9} & 92.5 & 31.1 & 51.1 & 65.6 & 51.0 & 38.3 & 19.7 & 26.7 & 60.6 & 0.0 & 24.5 & 32.8 & 40.5 & 0.0 & 24.9 & 59.3 & 27.1 & 47.2 & 22.7 & 39.3 \\
        \hline
        Tangent~\cite{tatarchenko2018tangent} & 63.3 & 91.8 & 36.9 & 64.6 & 64.5 & 56.2 & 42.7 & 27.9 & 35.2 & 47.4 & 14.7 & 35.3 & 28.2 & 25.8 & 28.3 & 29.4 & 61.9 & 48.7 & 43.7 & 29.8 & 43.8\\
        \hline
        3DMV~\cite{dai20183dmv} & 60.2 & 79.6 & 42.4 & 53.8 & 60.6 & 50.7 & 41.3 & 37.8 & 53.9 & 64.3 & 21.4 & 31.0 & 43.3 & 57.4 & \textbf{53.7} & 20.8 & 69.3 & 47.2 & 48.4 & 30.1 & 48.4\\
        \hline
        Ours & 68.0 & \textbf{93.5} & \textbf{49.4} & \textbf{66.4} & \textbf{71.9} & \textbf{63.6} & \textbf{46.4} & \textbf{39.6} & \textbf{56.8} & \textbf{67.1} & \textbf{22.5} & \textbf{44.5} & \textbf{41.1} & \textbf{67.8} & 41.2 & \textbf{53.5} & \textbf{79.4} & \textbf{56.5} & \textbf{67.2} & \textbf{35.6} & \textbf{56.6}\\
        \hline
    \end{tabular}
    (a) ScanNet (v2) (mean class IoU)
    \centering
    \tabcolsep=0.03cm
    \begin{tabular}{|c|c|c|c|c|c|c|c|c|c|c|c|c|c|c|c|c|c|c|c|c|c||c|}
        \hline
        Input & wall & floor & cab & bed & chair & sofa & table & door & wind & shf & pic & cntr & desk & curt & ceil & fridg & show & toil & sink & bath & other & avg\\
        \hline
        PN$^+$~\cite{qi2017pointnet++} & 80.1 & 81.3 & 34.1 & 71.8 & 59.7 & 63.5 & \textbf{58.1} & 49.6 & 28.7 & 1.1 & 34.3 & 10.1 & 0.0 & 68.8 & 79.3 & 0.0 & 29.0 & 70.4 & 29.4 & 62.1 & 8.5 & 43.8 \\
        \hline
        SplatNet~\cite{su2018splatnet} & \textbf{90.8} & \textbf{95.7} & 30.3 & 19.9 & \textbf{77.6} & 36.9 & 19.8 & 33.6 & 15.8 & 15.7 & 0.0 & 0.0 & 0.0 & 12.3 & 75.7 & 0.0 & 0.0 & 10.6 & 4.1 & 20.3 & 1.7 & 26.7 \\
        \hline
        Tangent~\cite{tatarchenko2018tangent} & 56.0 & 87.7 & 41.5 & 73.6 & 60.7 & 69.3 & 38.1 & 55.0 & 30.7 & 33.9 & 50.6 & 38.5 & 19.7 & 48.0 & 45.1 & 22.6 & 35.9 & 50.7 & 49.3 & 56.4 & 16.6 & 46.8 \\
        \hline
        3DMV~\cite{dai20183dmv} & 79.6 & 95.5 & \textbf{59.7} & 82.3 & 70.5 & \textbf{73.3} & 48.5 & 64.3 & 55.7 & 8.3 & 55.4 & 34.8 & 2.4 & \textbf{80.1} & \textbf{94.8} & 4.7 & 54.0 & 71.1 & 47.5 & 76.7 & 19.9 & 56.1 \\
        \hline
        Ours & 63.6 & 91.3 & 47.6 & \textbf{82.4} & 66.5 & 64.5 & 45.5 & \textbf{69.4} & \textbf{60.9} & \textbf{30.5} & \textbf{77.0} & \textbf{42.3} & \textbf{44.3} & 75.2 & 92.3 & \textbf{49.1} & \textbf{66.0} & \textbf{80.1} & \textbf{60.6} & \textbf{86.4} & \textbf{27.5} & \textbf{63.0} \\
        \hline
    \end{tabular}
    (b) Matterport3D (mean class accuracy)
    \vspace{-0.3cm}
    \caption{Comparison with the state-of-the-art methods for 3D semantic segmentation on the (a) ScanNet v2, and (b) Matterport3D~\cite{chang2017matterport3d} benchmarks. PN$^+$, SplatNet, and Tangent Convolution use points with per-point normal and color as input. 3DMV uses 2D images and voxels.  Ours uses grid points with high-res 10x10 texture patches.}
    \label{tab:mainresult}
\end{table*}

\begin{figure*}
    \centering
    \includegraphics[width=\linewidth]{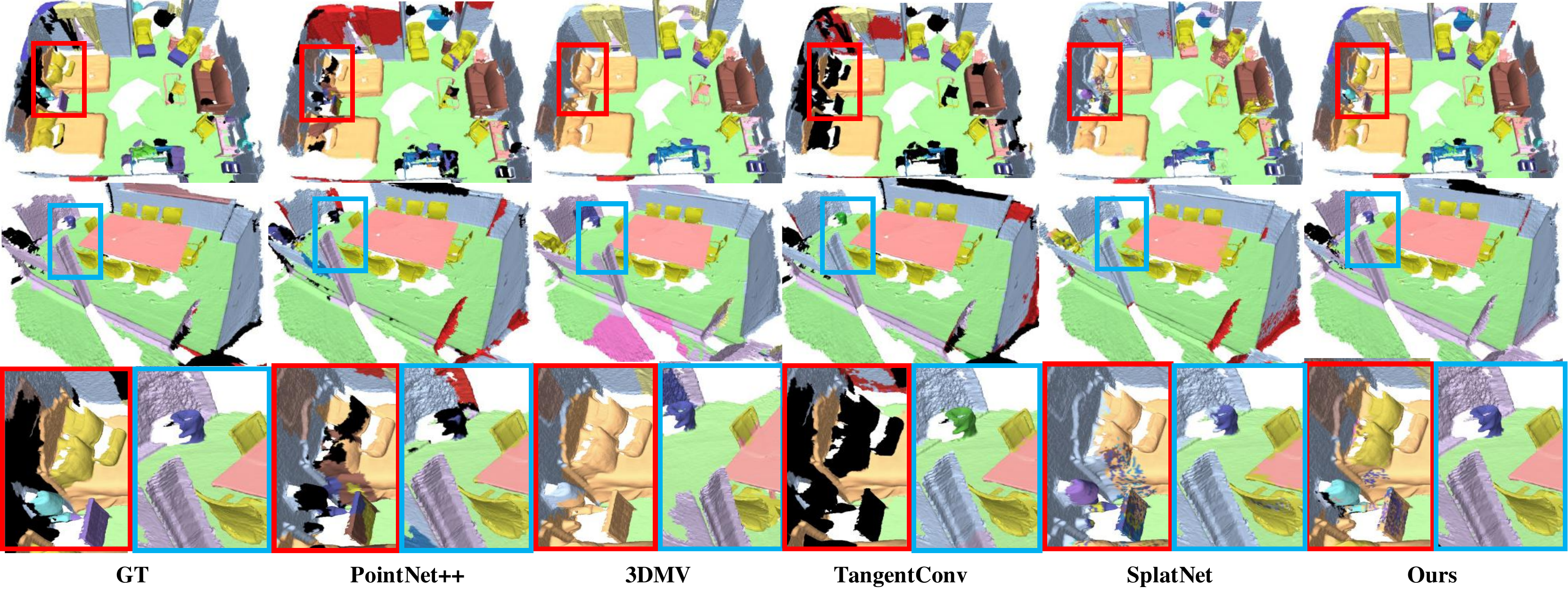}
    \vspace{-0.5cm}
    \caption{Visualization on ScanNet (v2)~\cite{dai2017scannet}. In the first row, we correctly predicts the lamp, pillow, picture, and part of the cabinet, while other methods fail. In the second row, we predict the window and the trash bin correctly, while 3DMV~\cite{dai20183dmv} predicts part of the window as the trash bin and other methods fail.  The third row (zoom-in) highlights the differences.}
    \label{fig:result-scannet}
    \vspace{-0.1in}
\end{figure*}

\begin{figure}[t]
    \centering
    \begin{minipage}{0.32\linewidth}
    \centering
    \includegraphics[width=\linewidth]{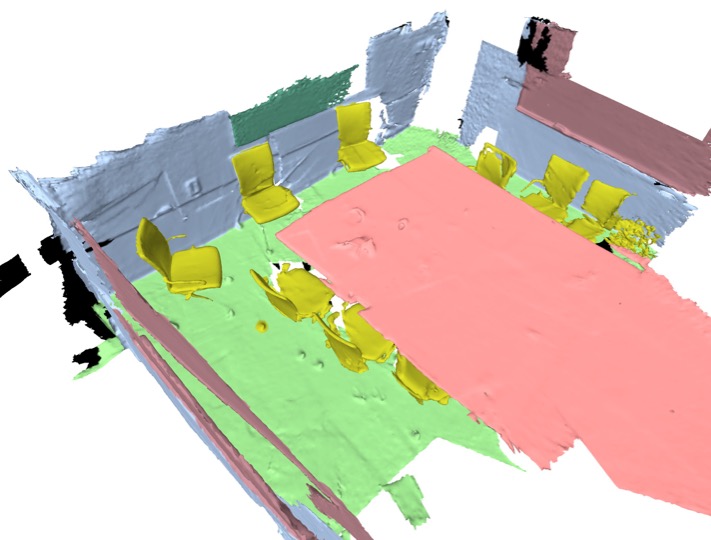}
    (a) Ground Truth
    \end{minipage}
    \begin{minipage}{0.32\linewidth}
    \centering
    \includegraphics[width=\linewidth]{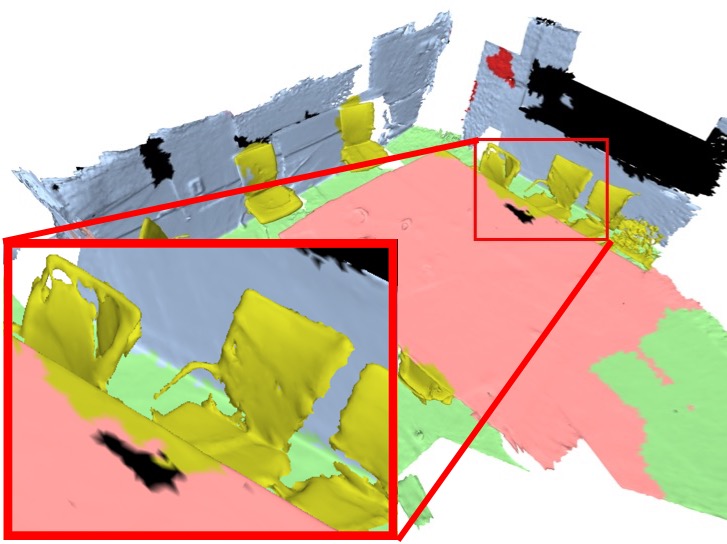}
    (b) Ball
    \end{minipage}
    \begin{minipage}{0.32\linewidth}
    \centering
    \includegraphics[width=\linewidth]{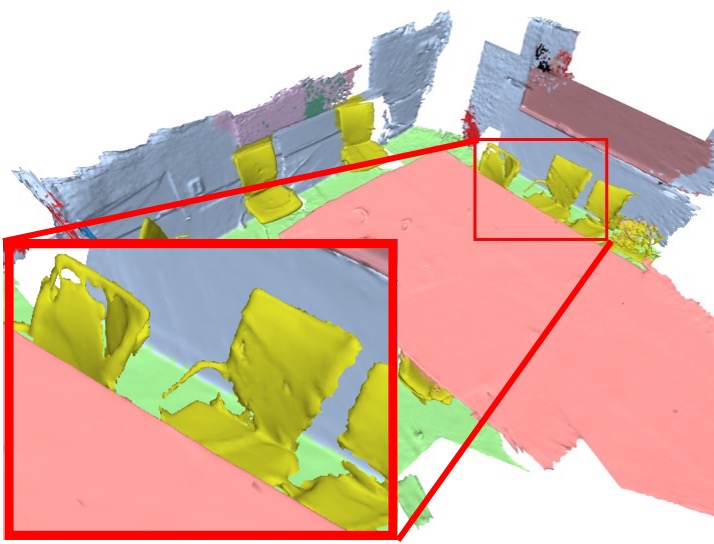}
    (c) Ours
    \end{minipage}
    \vspace{-0.2cm}
    \caption{Visual results using different neighborhoods. With euclidean ball as a neighborhood, part of the table is predicted as the chair, since they belong to the same euclidean ball. This issue is solved by extracting features from the geodesic patches.}
    \vspace{-0.2in}
    \label{fig:neighbor}
\end{figure}

\begin{figure}[t]
    \centering
    \includegraphics[width=\linewidth]{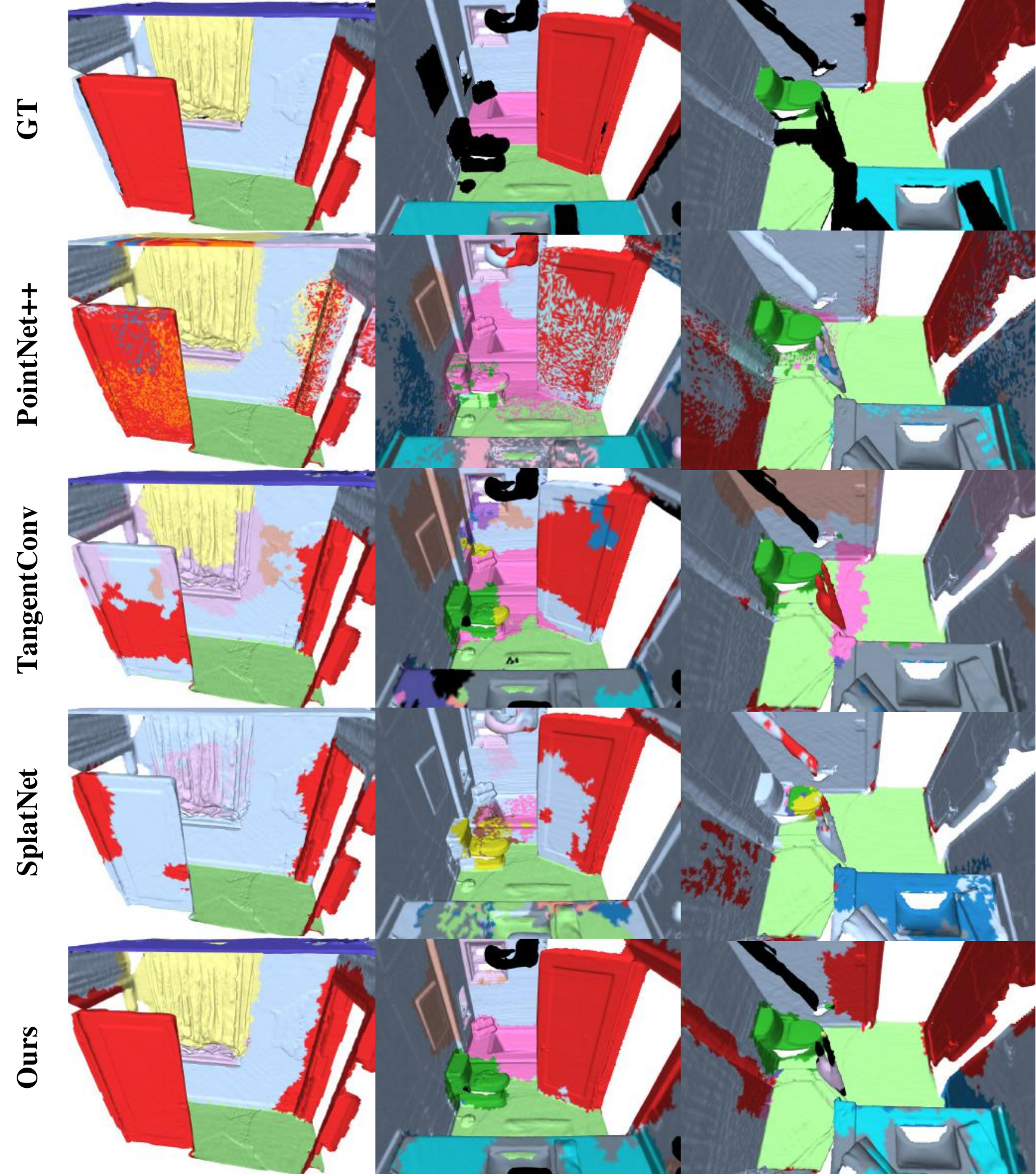}
    \vspace{-0.6cm}
    \caption{Visual results on Matterport3D~\cite{chang2017matterport3d}. In all examples, our method is better at predicting the door, the toilet, the sink, the bathtub, and the curtain.}
    \vspace{-0.2in}
    \label{fig:result-Matterport3D}
\end{figure}

Our main result is a comparison of \OURS{} to state-of-the-art methods for 3D semantic segmentation.  For this experiment, all methods utilize both color and geometry in their native formats.   Specifically, PointNet++~\cite{qi2017pointnet++}, Tangent Convolution~\cite{tatarchenko2018tangent}, SplatNet~\cite{su2018splatnet} use points with per-point normals and colors; 3DMV~\cite{dai20183dmv} uses 2D image features back-projected onto voxels; and Ours uses high-res 10x10 texture patches extracted from geodesic neighborhoods at sample points.

Table~\ref{tab:mainresult} reports the mean IoU scores for all 20 classes of the ScanNet benchmark on the ScanNet (v2) and mean class accuracy on Matterport3D datasets.   They show that \OURS{} (Ours) provides the best results on 18/20 classes for Scannet and 12/20 classes for Matterport3D.  Overall, the mean class IoU for Ours is 8.2\% higher than the previous state-of-the-art (3DMV) on ScanNet (48.4\% vs. 56.6\%), and our mean class accuracy is 6.9\% higher on Matterport3D (56.1\% vs. 63.0\%).  

Qualitative visual comparisons of the results shown in Figures~\ref{fig:result-scannet}-\ref{fig:result-Matterport3D} suggest that the differences between methods are often where high-resolution surface patterns are discriminating (e.g., the curtain and pillows in the top row of Figure~\ref{fig:result-scannet}) and where geodesic neighborhoods are more informative than Euclidean ones (e.g., the lamp next to the bed).  Figure~\ref{fig:neighbor} shows a case where convolutions with the geodesic neighborhoods clearly outperform their Euclidean counterparts. In Figure~\ref{fig:neighbor}(b), part of the table is predicted as chair, probably because it is in a Euclidean ball covering nearby chairs. This problem is solved with our method based on geodesic patch neighborhoods. As shown in Figure~\ref{fig:neighbor}(c), the table and the chairs are clearly segmented.

\para{Effect of 4-RoSy Surface Parameterization.}

\begin{figure}
    \centering
    \begin{minipage}{0.48\linewidth}
    \centering
    \includegraphics[width=\linewidth]{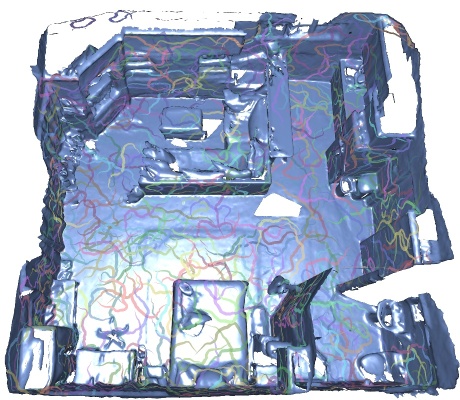}
    (a) RandomVec
    \end{minipage}
    \begin{minipage}{0.48\linewidth}
    \centering
    \includegraphics[width=\linewidth]{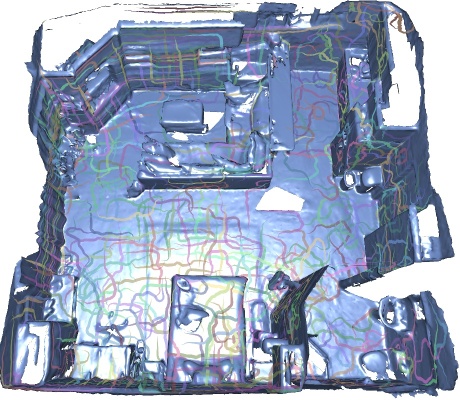}
    (b) EigenVec
    \end{minipage}
    \begin{minipage}{0.48\linewidth}
    \centering
    \includegraphics[width=\linewidth]{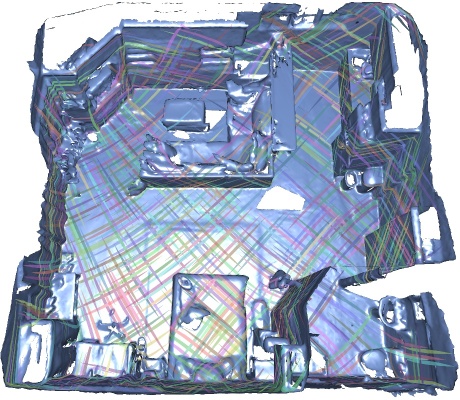}
    (c) Intrinsic
    \end{minipage}
    \begin{minipage}{0.48\linewidth}
    \centering
    \includegraphics[width=\linewidth]{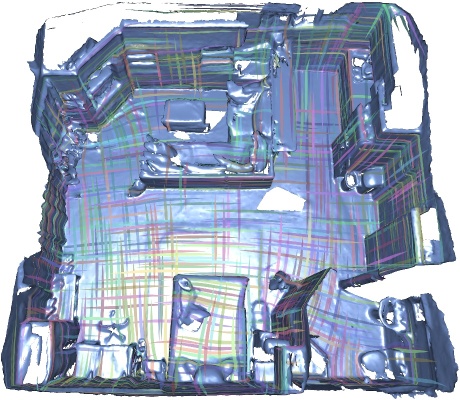}
    (d) Extrinsic
    \end{minipage}
    \vspace{-0.2cm}
    \caption{Direction fields from different methods. (a) Random directions lead to inconsistent frames. (b) Eigenvectors suffer from the same issue at flat area. (c) Intrinsic-energy based orientation field does not align to the shape features. (d) Our extrinsic-based method generates consistent orientation fields aligned with surface features.}
    \vspace{-0.2in}
    \label{fig:coordinate}
\end{figure}

Our second experiment is designed to test how different surface parameterizations affect semantic segmentation performance -- i.e., how does the choice of the orientation field affect the learning process?   The simplest choice is to pick an arbitrary direction on the tangent plane as the x-axis, similar to GCNN~\cite{masci2015geodesic}, (Figure~\ref{fig:coordinate}(a)).  A second option adopted by Tangent Convolution~\cite{tatarchenko2018tangent} considers a set of points $\mathbf{q}$ in a Euclidean ball centered at $\mathbf{p}$ and parameterizes the tangent plane by two eigenvectors corresponding to the largest two eigenvalues of the covariance matrix $\sum_{q}(p-q)(p-q)^T$.  A critical problem of this formulation is that the principal directions cannot be robustly analyzed at planar regions or noisy surfaces (Figure~\ref{fig:coordinate}(b)). It also introduces inconsistency to the coordinate systems of the neighboring points, which vexes the feature aggregation at higher levels.  A third alternative is to use the intrinsic energy function~\cite{jakob2015instant} or other widely used direction field synthesis technique~\cite{ray2008n,lai2010metric}, which is not geometry-aware and therefore variant to 3D rigid transformation (Figure~\ref{fig:coordinate}(c)). Our choice is to use the extrinsic energy to synthesize the direction field~\cite{huang2018quadriflow,jakob2015instant}, which is globally consistent and only variant to geometry itself (Figure~\ref{fig:coordinate}(d)).

\begin{table*}
    \centering
    \tabcolsep=0.055cm
    \begin{tabular}{|c|c|c|c|c|c|c|c|c|c|c|c|c|c|c|c|c|c|c|c|c|c|c|}
        \hline
        Input & wall & floor & cab & bed & chair & sofa & table & door & wind & bkshf & pic & cntr & desk & curt & fridg & show & toil & sink & bath & other & ave\\
        \hline
        Random & 37.6 & \textbf{92.5} & 37.0 & 63.7 & 28.5 & 56.9 & 27.6 & 15.3 & 31.0 & 47.6 & 16.5 & 36.6 & \textbf{53.3} & \textbf{51.2} & 15.4 & 24.7 & 59.3 & 47.6 & 53.3 & 27.0 & 41.1 \\
        \hline
        Intrinsic & 47.4 & 91.9 & 35.3 & 62.5 & 55.8 & 44.8 & 37.5 & 29.8 & 40.5 & 40.9 & 16.7 & 41.5 & 39.9 & 42.1 & 20.4 & 24.3 & 85.6 & 44.5 & 58.3 & 29.5 & 44.4 \\
        \hline
        EigenVec & 45.3 & 79.0 & 32.2 & 53.4 & 59.8 & 40.4 & 32.2 & 28.8 & 40.5 & 43.4 & \textbf{17.8} & 39.5 & 32.7 & 40.6 & 22.5 & 25.0 & 82.4 & 48.1 & 54.8 & 32.6 & 42.5 \\
        \hline
        Extrinsic & \textbf{69.8} & 92.3 & \textbf{44.8} & \textbf{69.4} & \textbf{75.8} & \textbf{67.1} & \textbf{56.8} & \textbf{39.4} & \textbf{41.1} & \textbf{63.1} & 15.8 & \textbf{57.4} & 46.5 & 48.3 & \textbf{36.9} & \textbf{40.0} & \textbf{78.1} & \textbf{54.0} & \textbf{65.4} & \textbf{34.4} & \textbf{54.8} \\
        \hline
    \end{tabular}
    \vspace{-0.3cm}
    \caption{Mean IoU for different direction fields on ScanNet (v2). The input is a pointcloud with a normal and rgb color for each point. {\em Random} refers to randomly picking an arbitrary direction for each sampled point. {\em Intrinsic} refers to solving for a 4-rosy field with intrinsic energy. {\em EigenVec} refers to solving for a direction field with the principal curvature. {\em Extrinsic} is our method, which solves a 4-rosy field with extrinsic energy.}
    \label{tab:direction}
    \vspace{-0.15in}
\end{table*}

To test the impact of this choice, we compare all of these alternative direction fields to create the local neighborhood parameterizations for our architecture and compare the results of 3D semantic segmentation on ScanNet (v1) test set.  As shown in Table~\ref{tab:direction}, the choice for random direction field performs worst since it does not provide consistent parameterization. The tangent convolution suffers from the same issue, but gets a better result since it aligns with the shape features.  The intrinsic parameterization aligns with the shape features, but is not a canonical parameterization -- for example, different rigid transformations of the same shape lead to different parameterizations. The extrinsic energy provides a canonical and consistent surface parameterization.  As a result, the extrinsic 4-rosy orientation field achieves the best results.

\begin{table*}
    \centering
    \tabcolsep=0.065cm
    \begin{tabular}{|c|c|c|c|c|c|c|c|c|c|c|c|c|c|c|c|c|c|c|c|c|c|}
        \hline
         Input & wall & floor & cab & bed & chair & sofa & table & door & wind & bkshf & pic & cntr & desk & curt & fridg & show & toil & sink & bath & other & ave\\
        \hline
         XYZ & 64.8 & 90.0 & 39.3 & 65.8 & 74.8 & 66.6 & 50.5 & 33.9 & 35.6 & 58.0 & 14.0 & 54.3 & 42.1 & 45.4 & 30.9 & 43.0 & 67.7 & 47.9 & 55.8 & 32.2 & 50.6 \\
        \hline
         NRGB & 69.8 & 92.3 & 44.8 & \textbf{69.4} & 75.8 & \textbf{67.1} & 56.8 & 39.4 & 41.1 & 63.1 & 15.8 & \textbf{57.4} & 46.5 & 48.3 & \textbf{36.9} & 40.0 & 78.1 & \textbf{54.0} & 65.4 & 34.4 & 54.8 \\
        \hline
         Highres & \textbf{75.0} & \textbf{94.4} & \textbf{46.8} & 67.3 & \textbf{78.1} & 64.0 & \textbf{63.5} & \textbf{44.8} & \textbf{46.0} & \textbf{71.3} & \textbf{21.1} & 44.4 & \textbf{47.5} & \textbf{52.5} & 35.2 & \textbf{51.3} & \textbf{80.3} & 51.7 & \textbf{67.6} & \textbf{40.2} & \textbf{58.1} \\
        \hline
    \end{tabular}
    \vspace{-0.3cm}
    \caption{Mean IoU for different color inputs on ScanNet (v2). {\em XYZ} represents our network using raw point input; i.e., geometry only. {\em NRGB} represents our network taking input as the sampled points with per-point normal and color. {\em Highres} represents our network taking per-point normal and the 10x10 surface texture patch for each sampled point.}
    \vspace{-0.3cm}
    \label{tab:highres}
\end{table*}

\para{Effect of 4-RoSy Surface Convolution.}

Our third experiment is designed to test how the choice for the surface convolution operator affects learning.  In Table~\ref{tab:operator}, PN$^+$(A) and PN$^+$ represent PointNet++ with average and max pooling, respectively. GCNN$^1$ and GCNN are geodesic convolutional neural networks~\cite{masci2015geodesic} with $N_\rho=3,N_\theta=1$ and $N_\rho=N_\theta=3$ respectively. ACNN represents anisotropic convolutional neural networks~\cite{boscaini2016learning} with $N_\rho=3,N_\theta=1$. RoSy$^1$ means a 3x3 convolution along the direction of the 1-rosy orientation field. RoSy$^4$ picks an arbitrary direction from the cross in the 4-rosy field. RoSy$^4$(m) applies 3x3 convolution for each direction of the cross in the 4-rosy field, aggregated by max pooling. Ours(A) and Ours represent our method with average and max pooling aggregation.

We find that GCNN, ACNN and RoSy$^4$ produce the lowest IoUs, because they suffer from inconsistency of frames when features are aggregated.  GCNN$^1$ does not suffer from this issue since there is only a single bin in the angle dimension. RoSy$^4$(m) uses the max-pooling to canonicalize the feature extraction, which is independent of the orientation selection, and produces better results than RoSy$^4$. RoSy$^1$ achieves a higher score by generating a more globally consistent orientation field with higher distortion. From this study, the combination of 4-rosy orientation field and our \OURS{} is the best option for the segmentation task among these methods. \jw{Since we precompute the local parametrization, our training efficiency is similar to that of GCNN.} Please refer to Supplemental~\ref{appendix:4rosy} for the detailed performance with each class.

\begin{table}
    \centering
    \tabcolsep=0.03cm
    \begin{tabular}{|c|c|c|c|c|c|}
        \hline
         Input & PN$^+$(A) & PN$^+$ & GCNN$^1$ & GCNN & ACNN\\
         \hline
         Geometry & 32.6 & 43.5 & 48.7 & 24.6 & 29.7\\
         \hline
         NRGB & 38.1 & 48.2 & 49.6 & 27.0 & 32.4\\
         \hline
         \multicolumn{6}{c}{}\\
         \hline
         Input &  RoSy$^1$ & RoSy$^4$ & RoSy$^1$(m) & Ours(A) & Ours\\
         \hline
         Geometry & 37.8 & 30.8 & 40.3 & 38.0 & \textbf{50.6}\\
         \hline
         NRGB & 47.8 & 34.5 & 42.6 & 39.1 & \textbf{54.8}\\
         \hline
    \end{tabular}
    \caption{Mean Class IoU with different texture convolution operators on ScanNet (v2). The input is the pointcloud for the first row (Geometry) and the pointcloud associated with the normal and rgb signal for the second row (NRGB).}
    \label{tab:operator}
    \vspace{-0.2in}
\end{table}

\para{Effect of High-Resolution Color.}

Our fourth experiment tests how much convolving with high-resolution surface colors affects semantic segmentation.   Table~\ref{tab:highres} compares the performance of our network with uncolored sampled points (XYZ), sampled points with the per-point surface normal and color (NRGB), and with the per-point normal and the 10x10 color texture patch (Highres) as input.  \jw{According to Table~\ref{tab:operator}, our network is already superior with only XYZ or additional NRGB because of the convolution operator.} We find that providing \OURS{} with Highres colors improves the mean class IoU by 3.3\%.  As expected, the impact is stronger from some semantic classes than others -- e.g., the IoUs for the bookshelf and picture classes increase 63.1$\rightarrow$71.3\% and 15.8$\rightarrow$21.1\%, respectively. \jw{We show additional comparison to O-CNN~\cite{wang2017cnn} which enables highres signals for voxels in Supplemental~\ref{appendix:ocnn}.}

\para{Comparisons Using Only Surface Geometry.}

As a final experiment, we evaluate the value of the proposed 3D network for semantic segmentation of inputs with only surface geometry (without color).  During experiments on ScanNet, \OURS{} achieves 50.6\% mIoU, which is 6.4\% better than the previous state-of-the-art.   In comparison, ScanNet~\cite{dai2017scannet} = 30.6\%, Tangent Convolution~\cite{tatarchenko2018tangent} = 40.9\%, PointNet++~\cite{qi2017pointnet++} = 43.5\%, and SplatNet~\cite{su2018splatnet} = 44.2\%.  Detailed class IoU results are provided in Supplemental~\ref{appendix:geometry}.

%% file: 5Conclusion.tex
\section{Conclusion}
\OURS{} bridges the gap between 2D image convolution and 3D deep learning using 4-RoSy surface parameterizations. We propose a new method for learning from high-resolution signals on 3D meshes by computing local geodesic neighborhoods with consistent 4-RoSy coordinate systems.  We design a network of 4-RoSy texture convolution operators that are able to learn surface features that significantly improve over the state-of-the-art performance for 3D semantic segmentation of 3D surfaces with color (by 6.9-8.2\%). Code and data will be publicly available.  Topics for further work include investigating the utility of \OURS{} for extracting features from other high-resolution signals on meshes (e.g., displacement maps, bump maps, curvature maps, etc.) and applications of \OURS{} to other computer vision tasks (e.g., instance detection, pose estimation, part decomposition, texture synthesis, etc.).

%% file: 6Appendix.tex
\section*{Supplemental}
\begin{appendix}
\section{Comparison to 2D Convolution on Texture Atlas}

\begin{figure}[h]
    \centering
    \includegraphics[width=\linewidth]{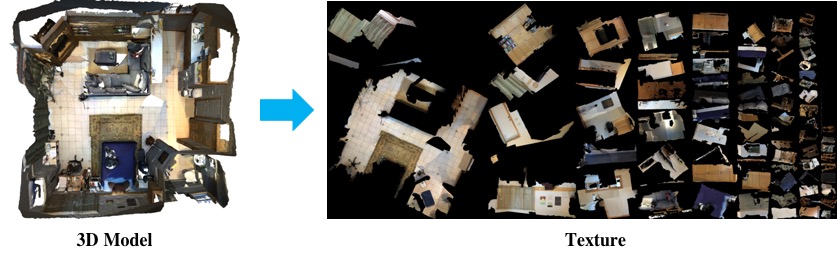}
    \caption{An example of the texture image.}
    \label{fig:textureimage}
\end{figure}

We did an additional experiment to compare our convolution operator with traditional image convolutions on a color texture atlas created with a standard UV parameterization, as shown in Figure~\ref{fig:textureimage}. For this experiment, we trained a state-of-the-art network (DenseNet~\cite{huang2017densely}) on the semantic labels mapped to the texture map image.  The results with that method are not very good -- the mean class IoU is only 12.2\%, as compared to 56.6\% with our method.  We conjecture the reason is that UV parameterizations are not consistent across examples and convolutions are affected by texture seams. 

\jw{We additionally tried an as-rigid-as-possible parameterization, which achieves 16.8 IoU (ours is 58.1).
The poor performance is mainly due to convolutions over regions with seams, large distortions, and inconsistent orientations -- i.e., the main problems that our 4-rosy approach aims to resolve.}

\section{Evaluation of Neighborhood Selection Methods}

The next experiment tests whether the geodesic neighborhoods used by \OURS{} convolutional operators are better than volumetric ones used by PointNet++.   To test this, we compare the performance of the original PointNet++ network which takes the Euclidean ball as the neighborhood, with slightly modified versions which take a cuboid or our geodesic patch as a neighborhood. As shown in Table~\ref{tab:neighbor}, the geodesic patch achieves a slightly higher score. This might be due to the reason that it is easier for the network to learn the boundary on the 2D subsurface than on the 3D space.

\begin{table*}[h]
    \centering
    \tabcolsep=0.068cm
    \begin{tabular}{|c|c|c|c|c|c|c|c|c|c|c|c|c|c|c|c|c|c|c|c|c|c|c|}
        \hline
        Input & wall & floor & cab & bed & chair & sofa & table & door & wind & bkshf & pic & cntr & desk & curt & fridg & show & toil & sink & bath & other & ave\\
        \hline
        Ball & \textbf{68.1} & \textbf{96.2} & 34.9 & 41.2 & 61.8 & 43.0 & 24.1 & 5.0 & 19.2 & 41.7 & 0.0 & 4.7 & 11.8 & 17.7 & 20.1 & 30.8 & 72.2 & 43.7 & 55.2 & 8.7 & 35.0 \\
        \hline
        Cube1 & 65.3 & 95.8 & 29.0 & 57.0 & 61.2 & 46.2 & 42.7 & 17.8 & 11.8 & 35.1 & 0.7 & \textbf{37.3} & \textbf{39.0} & 55.4 & 8.5 & 43.9 & 63.0 & 30.6 & 52.4 & 15.0 & 40.4 \\
        \hline
        Cube2 & 58.7 & 90.0 & \textbf{61.6} & \textbf{62.6} & 59.3 & 50.4 & 40.2 & 31.3 & 15.1 & 45.6 & 1.9 & 29.4 & 23.9 & 53.1 & 18.2 & 41.8 & 81.7 & 34.1 & 51.8 & 25.2 & 43.9 \\
        \hline
        Cube4 & 32.7 & 86.8 & 59.6 & 49.1 & 51.3 & 33.7 & 30.0 & 27.0 & 11.8 & 33.8 & 0.9 & 20.9 & 19.5 & 40.3 & 15.1 & 29.8 & 54.1 & 27.7 & 41.7 & 17.0 & 34.2 \\
        \hline
        Ours & 61.5 & 95.0 & 40.1 & 60.0 & \textbf{74.9} & \textbf{52.8} & \textbf{46.1} & \textbf{31.6} & \textbf{19.7} & \textbf{50.3} & \textbf{5.9} & 33.9 & 25.9 & \textbf{58.2} & \textbf{30.0} & \textbf{48.6} & \textbf{85.2} & \textbf{47.1} & \textbf{48.8} & \textbf{28.5} & \textbf{47.2} \\
        \hline
    \end{tabular}
    \caption{PointNet++ prediction using different neighborhood. The input is the sampled positions computed with our sampling method. Ball represents the euclidean ball. CubeX represents a tangent cuboid with the same volume as that of the ball, but has the width and length X times of the ball radius. Ours is using the geodesic patch with the same radius of the ball.}
    \label{tab:neighbor}
\end{table*}

\section{Effect of Point Sampling Method}
\label{sec:eval-sample}

\begin{table*}[h]
    \centering
    \tabcolsep=0.072cm
    \begin{tabular}{|c|c|c|c|c|c|c|c|c|c|c|c|c|c|c|c|c|c|c|c|c|c|c|}
        \hline
        Input & wall & floor & cab & bed & chair & sofa & table & door & wind & bkshf & pic & cntr & desk & curt & fridg & show & toil & sink & bath & other & ave\\
        \hline
        FPS & \textbf{70.2} & \textbf{92.3} & 43.1 & 63.7 & 67.7 & 62.5 & 50.8 & 23.4 & \textbf{42.5} & \textbf{65.2} & 15.4 & 54.7 & 44.3 & 45.0 & \textbf{40.1} & 33.5 & 71.6 & \textbf{54.3} & 62.4 & 28.7 & 51.6 \\
        \hline
         Quad & 69.8 & 92.3 & \textbf{44.8} & \textbf{69.4} & \textbf{75.8} & \textbf{67.1} & \textbf{56.8} & \textbf{39.4} & 41.1 & 63.1 & \textbf{15.8} & \textbf{57.4} & \textbf{46.5} & \textbf{48.3} & 36.9 & \textbf{40.0} & \textbf{78.1} & 54.0 & \textbf{65.4} & \textbf{34.4} & \textbf{54.8} \\
        \hline
    \end{tabular}
    \caption{PointNet++ prediction taking the positions of the pointcloud from different sampling methods including the furthest point sampling (FPS) and Quadriflow (Quad).}
    \label{tab:sample}
\end{table*}

\begin{figure}[h]
    \centering
    \begin{minipage}{0.48\linewidth}
    \centering
    \includegraphics[width=\linewidth]{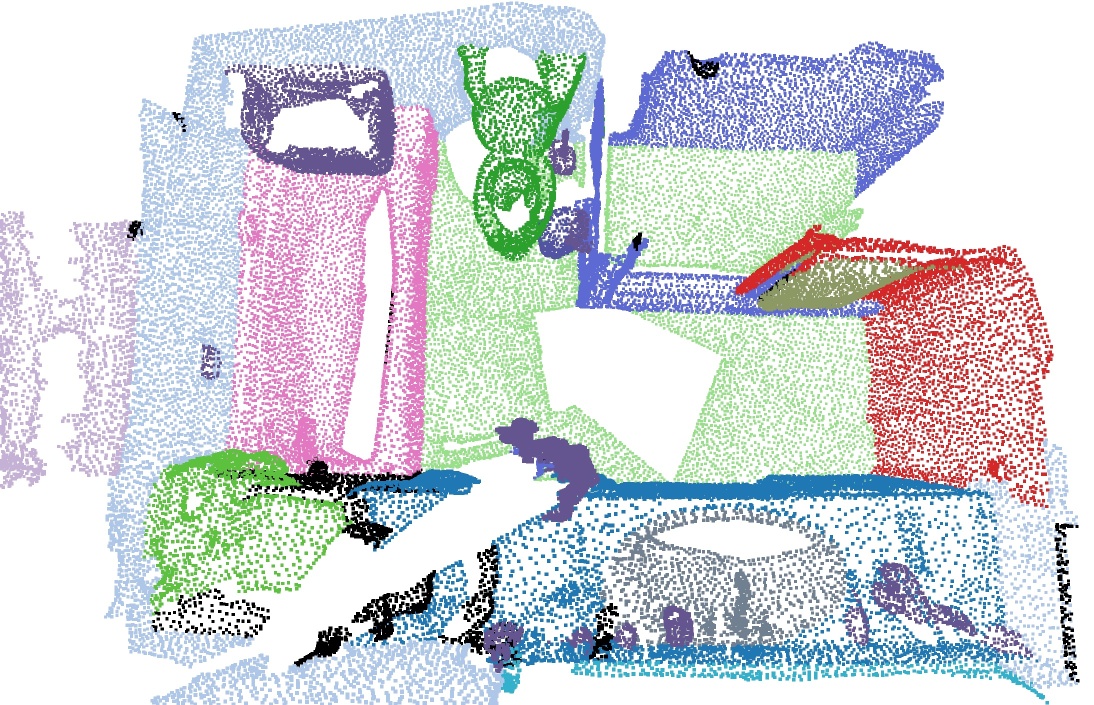}
    (a) Furthest Point Sampling
    \end{minipage}
    \begin{minipage}{0.48\linewidth}
    \centering
    \includegraphics[width=\linewidth]{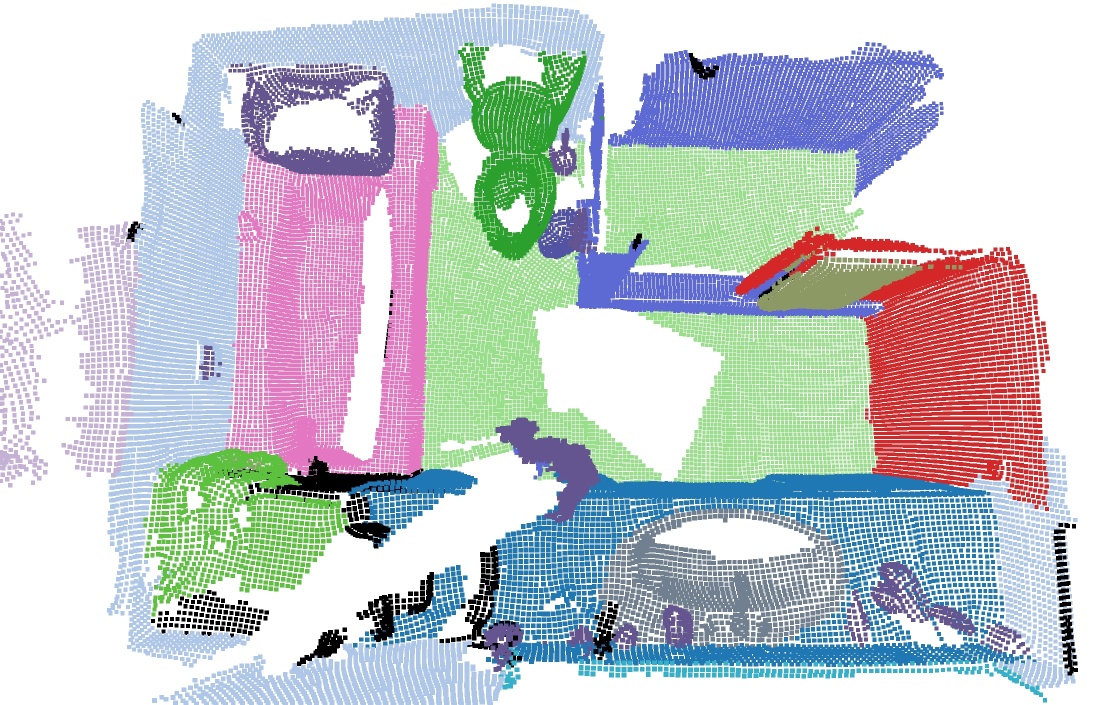}
    (b) Ours
    \end{minipage}\\
    \vspace{0.1cm}
    \caption{Visualization of Different Sampling methods.}
    \label{fig:sampling}
\end{figure}

The next experiment tests the impact of our proposed point sampling method.  While PointNet++~\cite{qi2017pointnet++} adopts the furthest point sampling method to preprocess the data, we use QuadriFlow~\cite{huang2018quadriflow} to sample the points on the surface. It maintains uniform edge length in surface parametrization, and therefore usually provides more uniformly distributed samples on the surface considering the geodesic distance. Figure~\ref{fig:sampling} shows the proportion of each class in the ScanNet dataset with QuadriFlow and furthest point sampling.

We use TextureNet to learn the semantic labels with their input and our samples. Table~\ref{tab:sample} shows the  class IoU for the prediction. With more samples for minor classes like the counter, desk, and curtain, our sampling method performs better. Figure~\ref{fig:sampling} shows the visualization of different sampling results. Visually, our sampling method leads to more uniformly distributed points on the surface.

\begin{figure}[h]
    \centering
    \includegraphics[width=\linewidth]{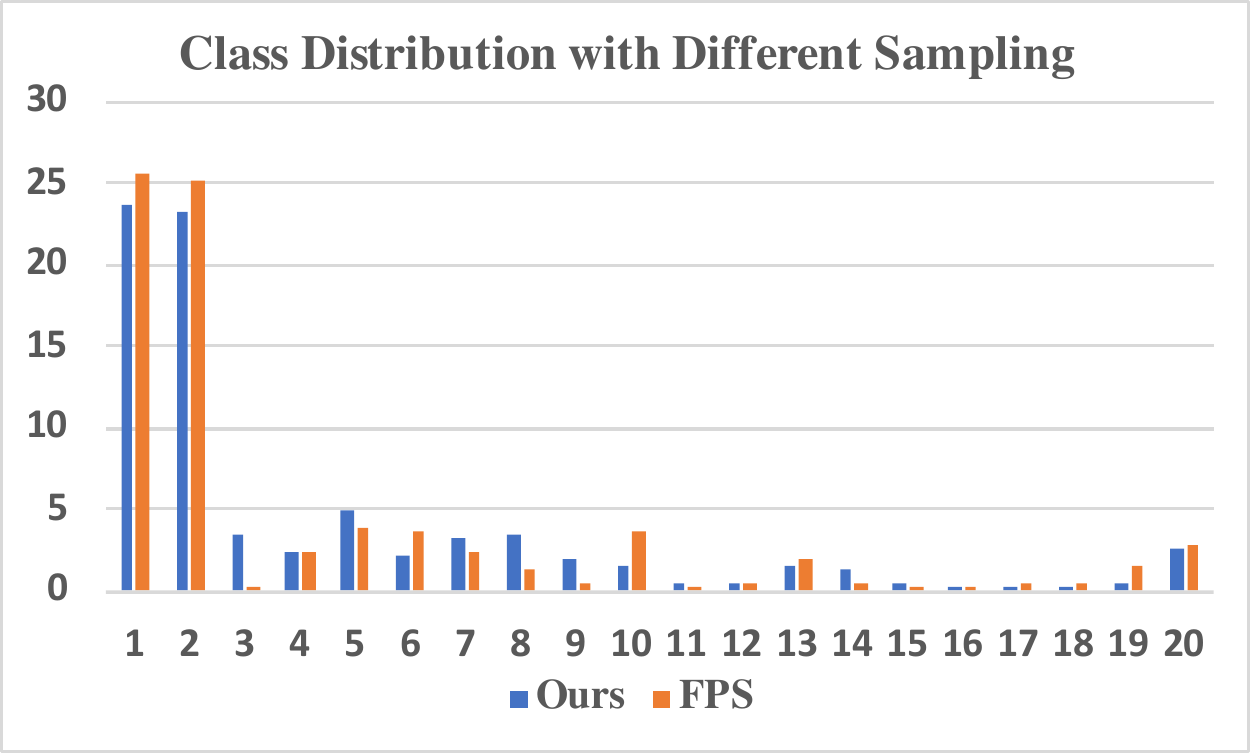}
    \caption{Class distribution with different sampling. The y-axis represents the portion of each class across all scenes. Except for classes of wall, floor, and bookshelf, our method achieves more samples than the furthest sampling method. As a result, PointNet++ achieves better results in most classes with our sampling method.}
    \label{fig:sampling-vis}
\end{figure}

\section{Further Results on Effect of 4-RoSy Surface Convolution}
\label{appendix:4rosy}
Table~\ref{tab:appendix-operator} provides detailed results for the performance of different surface convolution operators on ScanNet dataset~\cite{dai2017scannet} with input as the point cloud or the point cloud associated with the normal and RGB color for each point (expanding on Table 4 of the paper).    
PN$^+$(A) and PN$^+$ represent PointNet++ with average-pooling and maxpooling, respectively. GCNN$^1$ and GCNN are geodesic convolutional neural networks~\cite{masci2015geodesic} with $N_\rho=3,N_\theta=1$ and $N_\rho=N_\theta=3$ respectively. 
ACNN represents anisotropic convolutional neural networks~\cite{boscaini2016learning} with $N_\rho=3=N_\theta=3$. RoSy$^1$ refers to a 3x3 convolution along the direction of the 1-rosy orientation field. 
RoSy$^4$ picks an arbitrary direction from the cross in the 4-rosy field. RoSy$^4$(m) applies 3x3 convolution for each direction of the cross in the 4-rosy field, aggregated by max pooling. 
Ours(A) and Ours represent our method with average-pooling and max-pooling aggregation.
\begin{table*}
    \centering
    \tabcolsep=0.05cm
    \begin{tabular}{|c|c|c|c|c|c|c|c|c|c|c|c|c|c|c|c|c|c|c|c|c|c|}
        \hline
         Operator & wall & floor & cab & bed & chair & sofa & table & door & wind & bkshf & pic & cntr & desk & curt & fridg & show & toil & sink & bath & other & ave\\
        \hline
         PN$^+$(A) & 55.7 & 80.2 & 23.1 & 41.6 & 54.1 & 55.9 & \textbf{68.6} & 11.2 & 20.0 & 41.1 & 5.3 & 37.5 & 36.2 & 4.7 & 2.9 & 6.0 & 30.6 & 21.9 & 48.1 & 7.8 & 32.6\\
         \hline
         PN$^+$ & \textbf{68.7} & 89.9 & 38.3 & 60.1 & 73.5 & 62.0 & 62.2 & 30.9 & 28.2 & 52.8 & 9.6 & 42.7 & 38.6 & 38.4 & 23.4 & 35.7 & 66.2 & 47.6 & 57.4 & 26.0 & 47.6\\
        \hline
         GCNN$^1$ & 62.5 & 94.0 & 35.8 & 65.6 & 73.2 & 63.9 & 59.5 & 30.0 & 32.0 & 57.6 & 11.6 & 53.0 & 38.9 & 40.6 & 29.7 & 46.0 & 59.8 & 43.8 & 48.9 & 27.5 & 48.7\\
        \hline
         GCNN & 54.4 & 81.8 & 17.4 & 9.9 & 48.1 & 24.2 & 28.2 & 16.0 & 24.5 & 15.5 & 9.5 & 18.8 & 15.1 & 27.7 & 6.3 & 20.3 & 27.7 & 23.0 & 9.1 & 13.5 & 24.6 \\
        \hline
         ACNN & 65.1 & 88.0 & 17.0 & 23.0 & 54.2 & 18.7 & 35.9 & 16.4 & 28.1 & 0.3 & \textbf{14.6} & 22.0 & 23.4 & 25.6 & 7.0 & 23.1 & 43.6 & 36.9 & 33.6 & 17.5 & 29.7 \\
        \hline
         RoSy$^1$ & 49.4 & 80.5 & 24.5 & 41.3 & 65.7 & 48.8 & 39.1 & 19.3 & 28.2 & 44.7 & 8.6 & 36.1 & 25.2 & 30.9 & 16.7 & 38.9 & 52.9 & 37.8 & 47.3 & 19.4 & 37.8 \\
         \hline
         RoSy$^4$ & 55.4 & 90.8 & 25.3 & 24.5 & 56.0 & 29.5 & 43.0 & 16.9 & 19.9 & 29.7 & 6.0 & 21.6 & 17.3 & 32.7 & 9.0 & 33.0 & 29.7 & 21.1 & 34.2 & 20.5 & 30.9 \\
         \hline
         RoSy$^4$(m) & 61.3 & 88.2 & 26.7 & 47.6 & \textbf{80.6} & 50.5 & 52.1 & 12.7 & 31.5 & 46.1 & 13.7 & 47.4 & 25.1 & 20.9 & 9.8 & 29.8 & 50.2 & 41.1 & 43.6 & 27.7 & 40.3 \\
         \hline
         Ours(A) & 51.5 & 87.1 & 26.0 & 44.7 & 65.0 & 46.4 & 42.5 & 18.5 & 31.4 & 29.0 & 8.0 & 40.6 & 24.9 & 11.5 & 18.9 & 34.9 & 61.2 & 43.0 & 50.2 & 23.8 & 38.0\\
        \hline
         Ours & 64.8 & \textbf{90.0} & \textbf{39.3} & \textbf{65.8} & 74.8 & \textbf{66.6} & 50.5 & \textbf{33.9} & \textbf{35.6} & \textbf{58.0} & 14.0 & \textbf{54.3} & \textbf{42.1} & \textbf{45.4} & \textbf{30.9} & \textbf{43.0} & \textbf{67.7} & \textbf{47.9} & \textbf{55.8} & \textbf{32.2} & \textbf{50.6} \\
        \hline
    \end{tabular}\\
    (a) Pointcloud\\
    \centering
    \tabcolsep=0.05cm
    \begin{tabular}{|c|c|c|c|c|c|c|c|c|c|c|c|c|c|c|c|c|c|c|c|c|c|}
        \hline
         Operator & wall & floor & cab & bed & chair & sofa & table & door & wind & bkshf & pic & cntr & desk & curt & fridg & show & toil & sink & bath & other & ave\\
        \hline
         PN$^+$(A) & 66.6 & 94.7 & 29.9 & 50.5 & 64.9 & 52.9 & 56.5 & 17.4 & 19.7 & 45.0 & 0.0 & 36.5 & 30.4 & 21.5 & 13.5 & 19.1 & 49.6 & 30.3 & 45.6 & 16.6 & 38.1\\
         \hline
         PN$^+$~\cite{qi2017pointnet++} & \textbf{81.5} & \textbf{95.0} & 40.1 & 60.0 & 74.9 & 52.8 & 46.1 & 31.3 & 19.7 & 50.3 & 5.9 & 33.9 & 25.9 & \textbf{58.2} & 30.0 & \textbf{48.6} & \textbf{85.2} & 47.1 & 48.8 & 28.5 & 48.2\\
        \hline
         GCNN$^1$ & 69.4 & 93.1 & 37.3 & 65.4 & 68.6 & 54.3 & \textbf{59.0} & 35.7 & 34.6 & 56.7 & 17.5 & 51.8 & 40.2 & 39.6 & 27.0 & 47.0 & 57.7 & 39.9 & \textbf{69.4} & 28.6 & 49.6 \\
        \hline
         GCNN~\cite{masci2015geodesic} & 46.8 & 89.1 & 21.1 & 31.5 & 52.1 & 36.6 & 41.6 & 17.2 & 18.1 & 21.3 & 3.7 & 23.5 & 17.7 & 22.6 & 4.9 & 16.7 & 24.6 & 22.7 & 16.9 & 11.3 & 27.0 \\
        \hline
         ACNN~\cite{boscaini2016anisotropic} & 58.4 & 89.2 & 23.8 & 30.6 & 61.5 & 29.7 & 39.4 & 18.5 & 25.4 & 14.2 & 5.1 & 33.7 & 19.2 & 29.0 & 8.6 & 30.7 & 41.6 & 35.5 & 36.4 & 17.0 & 32.4 \\
        \hline
         RoSy$^1$ & 56.3 & 90.9 & 34.9 & 50.5 & 73.5 & 58.6 & 51.7 & 30.7 & 39.9 & 56.1 & 9.7 & 45.1 & 36.7 & 39.5 & 28.2 & 42.8 & 68.6 & 49.5 & 64.6 & 29.3 & 47.8 \\
         \hline
         RoSy$^4$ & 51.7 & 89.3 & 26.0 & 39.1 & 60.8 & 37.4 & 42.8 & 10.4 & 30.6 & 39.1 & 14.9 & 35.9 & 19.7 & 17.4 & 8.6 & 21.0 & 42.3 & 38.0 & 36.4 & 19.6 & 34.5 \\
         \hline
         RoSy$^4$(m) & 66.2 & 93.4 & 33.7 & 50.3 & \textbf{78.5} & 47.6 & 54.9 & 13.4 & 39.0 & 49.7 & \textbf{18.8} & 46.5 & 24.9 & 22.2 & 10.7 & 27.2 & 54.2 & 48.8 & 46.5 & 25.4 & 42.6 \\
         \hline
         Ours(A) & 52.4 & 91.3 & 29.1 & 42.5 & 65.6 & 42.1 & 47.3 & 20.6 & 31.4 & 30.9 & 7.3 & 40.8 & 26.2 & 10.7 & 18.2 & 31.2 & 64.8 & 44.1 & 63.6 & 21.1 & 39.1 \\
        \hline
         Ours & 69.8 & 92.3 & \textbf{44.8} & \textbf{69.4} & 75.8 & \textbf{67.1} & 56.8 & \textbf{39.4} & \textbf{41.1} & \textbf{63.1} & 15.8 & \textbf{57.4} & \textbf{46.5} & 48.3 & \textbf{36.9} & 40.0 & 78.1 & \textbf{54.0} & 65.4 & \textbf{34.4} & \textbf{54.8} \\
        \hline
    \end{tabular}
    (b) Pointcloud with per-point normal and RGB color\\
    \caption{Texture Convolution Operator Comparison. The input is the pointcloud in (a) and the pointcloud associated with the normal and the rgb color for each point in (b). PN$^+$(A) and PN$^+$ represent PointNet++ with average-pooling and maxpooling, respectively. GCNN$^1$ and GCNN are geodesic convolutional neural networks~\cite{masci2015geodesic} with $N_\rho=3,N_\theta=1$ and $N_\rho=N_\theta=3$ respectively. ACNN represents anisotropic convolutional neural networks~\cite{boscaini2016learning} with $N_\rho=3,N_\theta=1$. RoSy$^1$ means a 3x3 convolution along the direction of the 1-rosy orientation field. RoSy$^4$ picks an arbitrary direction from the cross in the 4-rosy field. RoSy$^4$(m) applies 3x3 convolution for each direction of the cross in the 4-rosy field, aggregated by maxpooling. Ours(A) and Ours represent our method with average-pooling and max-pooling aggregation.}
    \label{tab:appendix-operator}
\end{table*}

\section{Comparison to Octree-based Approaches} 
\label{appendix:ocnn}
\jw{Existing volume-based octree methods have been used mostly for stand-alone objects from ShapeNet.
For larger scenes, memory is a severe limitation.
As a test, we tried O-CNN~\cite{wang2017cnn} on chunks of radius 1.35m$^3$ using a 12GB GPU, which fits 6 conv/deconv layers and a feature dimension of 256 at resolution $256^3$.  This test yielded a mean IoU of 30.8 with NRGB and 27.8 with pure geometry. 
In contrast, the surface-based convolution of TextureNet is much more efficient (2D rather than 3D), allowing for a total of 18 conv/deconv layers with max feature dimension of 1024, and achieves 58.1 with high-res color, 54.8 with NRGB, and 50.6 with pure geometry.  }

\section{Further Comparisons Using Only Surface Geometry}
\label{appendix:geometry}
This section provides more detailed results for the experiment described in the last paragraph of Section 4 of the paper, where we evaluate the value of the proposed 3D network for semantic segmentation of inputs with only surface geometry (without color).  During experiments on ScanNet, \OURS{} achieves 50.6\% mIoU, which is 6.4\% better than the previous state-of-the-art.   In comparison, ScanNet~\cite{dai2017scannet} = 30.6\%, Tangent Convolution~\cite{tatarchenko2018tangent} = 40.9\%, PointNet++~\cite{qi2017pointnet++} = 43.5\%, and SplatNet~\cite{su2018splatnet} = 44.2\%.  Detailed class IoU results are provided in Table~\ref{tab:geometry}.

\begin{table*}
    \centering
    \tabcolsep=0.05cm
    \begin{tabular}{|c|c|c|c|c|c|c|c|c|c|c|c|c|c|c|c|c|c|c|c|c|c|}
        \hline
        Input & wall & floor & cab & bed & chair & sofa & table & door & wind & shf & pic & cntr & desk & curt & fridg & show & toil & sink & bath & other & avg\\
        \hline
        ScanNet~\cite{dai2017scannet} & 43.7 & 78.6 & 31.1 & 36.6 & 52.4 & 34.8 & 30.0 & 18.9 & 18.2 & 50.1 & 10.2 & 21.1 & 34.2 & 0.0 & 24.5 & 15.2 & 46.0 & 31.8 & 20.3 & 14.5 & 30.6\\
        \hline
        PN$^+$~\cite{qi2017pointnet++} & 64.1 & 82.2 & 31.4 & 51.6 & 64.5 & 51.4 & 44.9 & 23.3 & 30.4 & \textbf{68.2} & 3.7 & 26.2 & 34.2 & \textbf{65.1} & 23.4 & 18.3 & 61.8 & 31.5 & \textbf{75.4} & 18.8 & 43.5\\
        \hline
        SplatNet~\cite{su2018splatnet} & \textbf{67.4} & 85.8 & 32.3 & 45.1 & 71.9 & 51.0 & 40.7 & 15.1 & 25.2 & 62.3 & 0.0 & 23.2 & 39.9 & 56.1 & 0.0 & 24.2 & 62.6 & 23 & 67.4 & 25.7 & 40.9\\
        \hline
        Tangent~\cite{tatarchenko2018tangent} & 62.0 & 83.6 & 39.3 & 58.4 & 67.6 & 57.3 & 47.9 & 27.6 & 28.5 & 55.0 & 8.3 & 36.1 & 33.9 & 38.7 & 26.2 & 28.0 & 60.5 & 39.3 & 59.0 & 27.8 & 44.2\\
        \hline
        Ours & 64.8 & \textbf{90.0} & \textbf{39.3} & \textbf{65.8} & \textbf{74.8} & \textbf{66.6} & \textbf{50.5} & \textbf{33.9} & \textbf{35.6} & 58.0 & \textbf{14.0} & \textbf{54.3} & \textbf{42.1} & 45.4 & \textbf{30.9} & \textbf{43.0} & \textbf{67.7} & \textbf{47.9} & 55.8 & \textbf{32.2} & \textbf{50.6}\\
        \hline
    \end{tabular}
    \caption{Geometry-only: comparison to the state-of-the-art for 3D convolution with pure geometry as input; i.e., no RGB information used in any of these experiments. We can show that our method also outperforms existing geometry-only approaches.}
    \label{tab:geometry}
\end{table*}

\section{Effect of 4-RoSy convolution on traditional image convolution}

We also compared our 4-RoSy operator with the traditional image convolution on the MNIST dataset~\cite{lecun2010mnist}. We use a simple network containing two MLP layers and two fully connected layers. The performance of the original network is 99.1\%. By replacing the convolution with our 4-RoSy operator in the MLP layers, we achieve 98.5\% classification accuracy. Therefore, our 4-RoSy kernel is comparable to the traditional convolutions even on the standard images.

\section{Visual comparison of Different Resolutions}
\begin{figure*}[t]
    \centering
    \includegraphics[width=0.8\textwidth]{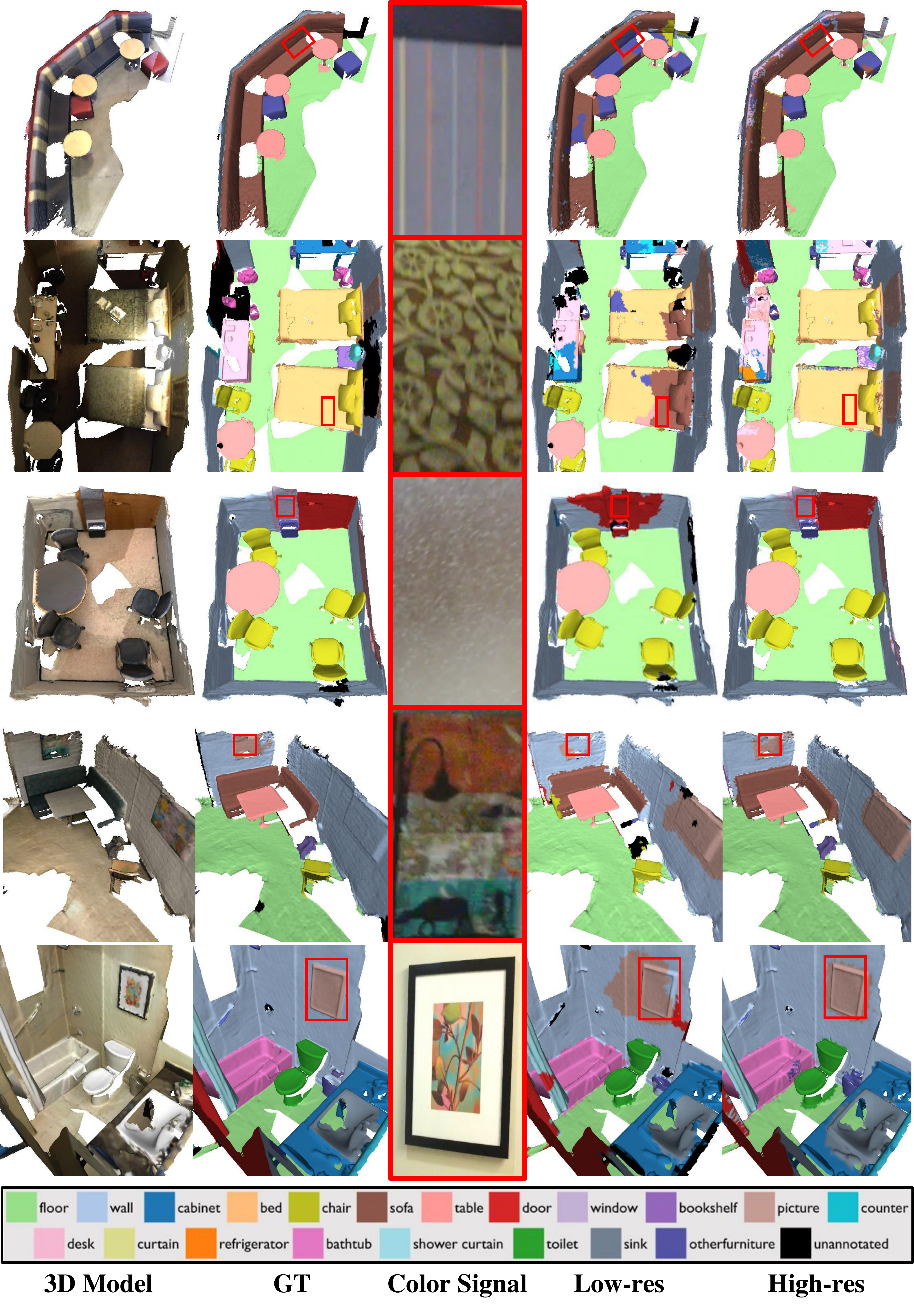}
    \caption{Visual comparison of Different Resolutions. The column row is the 3D model. The second column shows the ground truth semantic labels. The high-res signals of the red regions are shown in the third column. The last two columns are predictions from TextureNet with per-point color (low-res) or high-res texture patch as input.}
    \label{fig:highres}
\end{figure*}

In Figure~\ref{fig:highres}, we show the predictions of TextureNet with different color resolutions as input. The first column is the 3D model. The second column shows the ground truth semantic labels. The high-res signals of the red regions are shown in the third column. The last two columns are predictions from TextureNet with per-point color (low-res) or high-res texture patch as input. As a result, TextureNet performs better given the input with high-res signals.

\section{Visualization of the Semantic Segmentation}
\begin{figure*}[p]
\centering
\includegraphics[width=0.75\textwidth]{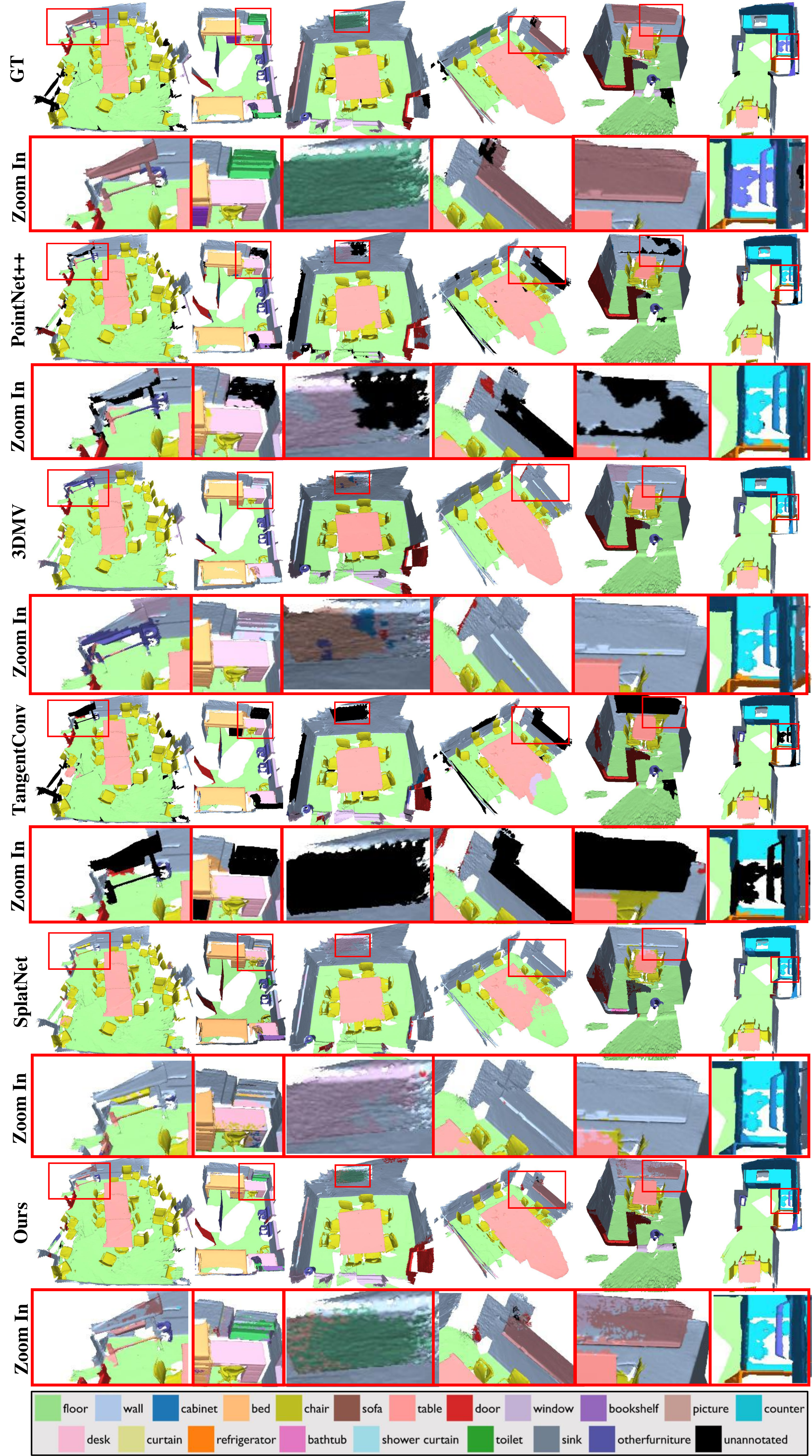}
\caption{Visualization of the Semantic Segmentation on ScanNet Dataset.}
\label{fig:supple-scannet1}
\end{figure*}

\begin{figure*}[p]
\centering
\includegraphics[width=0.8\textwidth]{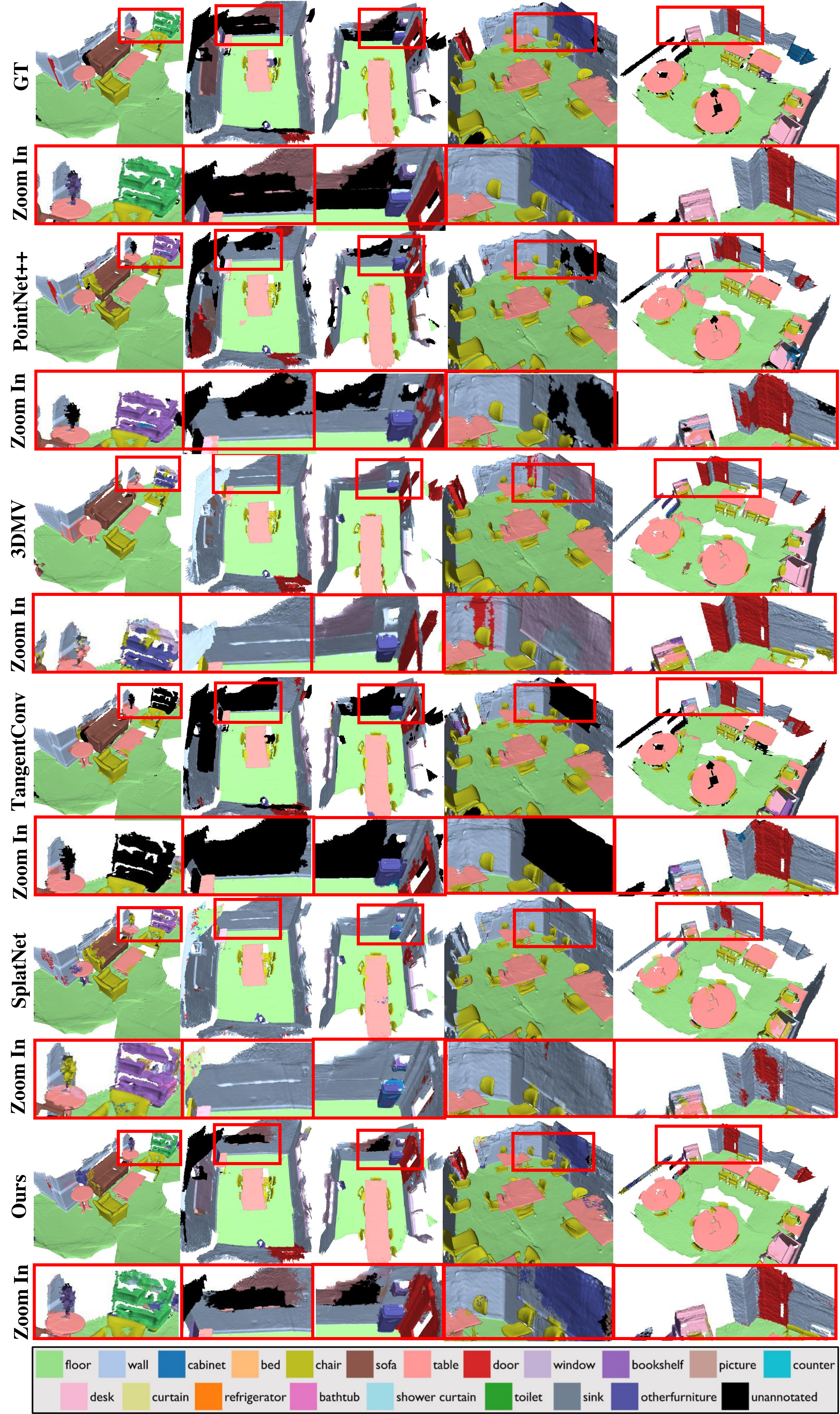}
\caption{Visualization of the Semantic Segmentation on ScanNet Dataset.}
\label{fig:supple-scannet2}
\end{figure*}

\begin{figure*}[p]
\centering
\includegraphics[width=0.8\textwidth]{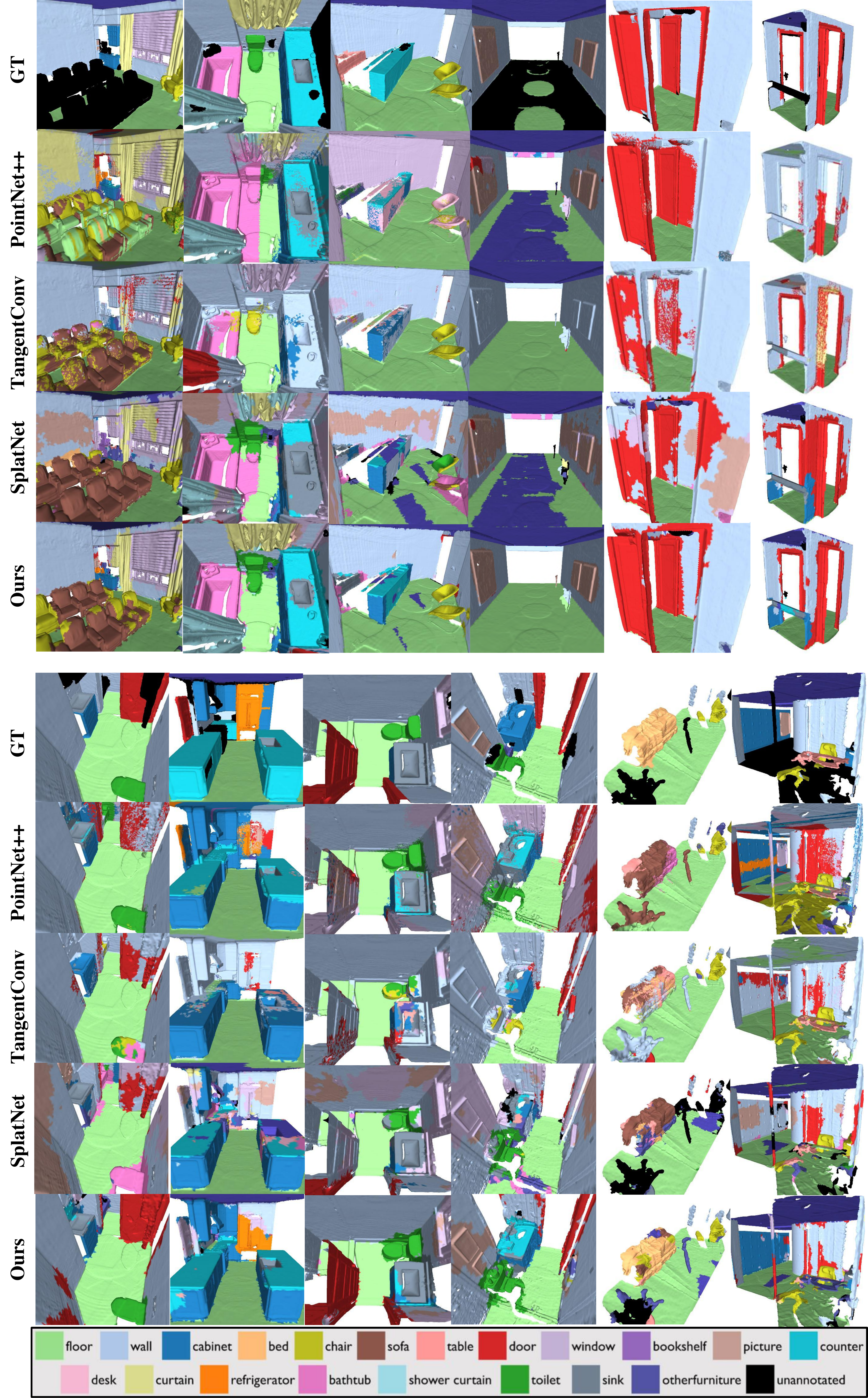}
\caption{Visualization of the Semantic Segmentation on Matterport Dataset.}
\label{fig:supple-matterport}
\end{figure*}

We compare TextureNet with the state-of-the-art method on ScanNet Dataset~\cite{dai2017scannet} and Matterport3D Dataset. On both datasets, we outperform existing methods (see the main paper). Figure~\ref{fig:supple-scannet1} and~\ref{fig:supple-scannet2} show examples of prediction from several methods on ScanNet. Figure~\ref{fig:supple-matterport} show examples of prediction from different methods on Matterport3D Dataset.

\end{appendix}